\PassOptionsToPackage{dvipsnames}{xcolor}
\documentclass[acmsmall,nonacm,screen,11pt]{acmart}
\acmYear{2026}
\usepackage{fontspec,hyperref}
\newfontface\ArxivBundledFontOne{AppliedSansDisplay-Bold.ttf}[NFSSFamily=ArxivBundledFontOne]
\newfontface\ArxivBundledFontTwo{AppliedSansDisplay-SemiBold.ttf}[NFSSFamily=ArxivBundledFontTwo]
\newfontface\ArxivBundledFontThree{AppliedSansText-Italic.ttf}[NFSSFamily=ArxivBundledFontThree]
\newfontface\ArxivBundledFontFour{AppliedSansText-Regular.ttf}[NFSSFamily=ArxivBundledFontFour]
\newfontface\ArxivBundledFontFive{AppliedSansText-SemiBold.ttf}[NFSSFamily=ArxivBundledFontFive]
\newfontface\ArxivBundledFontSix{AppliedSansText-SemiBoldItalic.ttf}[NFSSFamily=ArxivBundledFontSix]
\newfontface\ArxivBundledFontSeven{XCharter-Bold.otf}[NFSSFamily=ArxivBundledFontSeven]
\newfontface\ArxivBundledFontEight{XCharter-BoldItalic.otf}[NFSSFamily=ArxivBundledFontEight]
\newfontface\ArxivBundledFontNine{XCharter-Italic.otf}[NFSSFamily=ArxivBundledFontNine]
\newfontface\ArxivBundledFontTen{XCharter-Roman.otf}[NFSSFamily=ArxivBundledFontTen]

\usepackage[layout=paper]{appliedpaper}
\usepackage{makecell,multirow,wrapfig,placeins,tabularx,array,bbm,enumitem,tikz}

\providecommand{\bm}[1]{\symbfit{#1}}

\def\eqref#1{equation~\ref{#1}}

\def\1{\bm{1}}

\newcommand{\cmark}{\ding{51}}
\newcommand{\xmark}{\ding{55}}
\renewcommand{\tabularxcolumn}[1]{m{#1}}
\definecolor{blockblue}{HTML}{BEEEFF}
\definecolor{blockorange}{HTML}{FFE3A2}
\definecolor{blockgrey}{HTML}{B3B3B3}
\usetikzlibrary{arrows.meta,patterns.meta}

\usetikzlibrary{patterns.meta}

\tikzset{
  symbolbox/.style={
    draw=black!60,
    fill=black!3,
    text=black,
    rounded corners=1.8pt,
    line width=0.8pt,
    minimum size=1.4em,
    inner sep=0pt,
    outer sep=0pt
  },
  symbolhatch/.style={
    pattern={
      Lines[
        angle=45,
        distance=2.5pt,
        line width=0.4pt
      ]
    },
    pattern color=black!60
  }
}

\definecolor{hatchgold}{HTML}{D99C00}
\definecolor{hatchcream}{HTML}{FFEACB}
\definecolor{hatchblue}{HTML}{6D91C2}
\definecolor{hatchlightblue}{HTML}{D8EBFF}

\setpaperteaser{\centering
    \includegraphics[width=\linewidth]{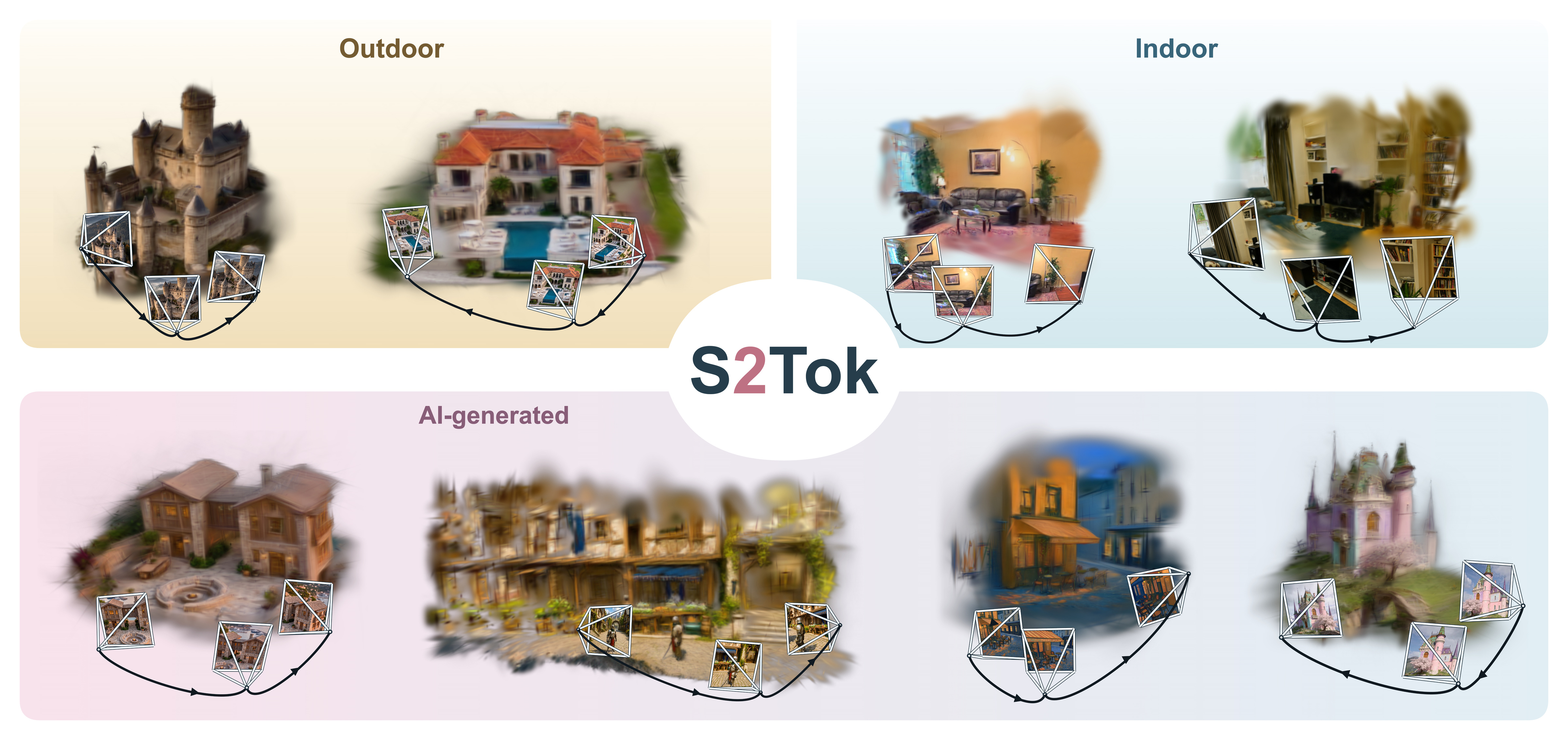}
    \caption{\textbf{S2Tok on Various Scenarios.} We show the final 3DGS scenes reconstructed by S2Tok in a streaming manner across diverse scenarios. More streaming reconstruction samples are in Fig.~\ref{fig:gsscene-mipnerf}.}
    \label{fig:teasor}
}

\begin{document}
\title[S2Tok: Streaming 3D Gaussian Reconstruction]{S2Tok: Streaming 3D Gaussian Reconstruction with Persistent Spatial Tokens}
\author{Fang Li}
\author{Jiraphon Yenphraphai}
\author{Quentin Herau}
\author{Depu Meng}
\author{Yihan Hu}
\author{Tianshuo Xu}
\author{Narendra Ahuja}
\author{Wei Zhan}
\compactauthors{%
  {\raggedright
   {\rmfamily\bfseries
    \appliedname{Fang Li\textsuperscript{1,2,*}}, \;
    \appliedname{Jiraphon Yenphraphai\textsuperscript{1,3,*}}, \;
    \appliedname{Quentin Herau\textsuperscript{1}}, \;
    \appliedname{Depu Meng\textsuperscript{1}},\\[0.25em]
    \appliedname{Yihan Hu\textsuperscript{1}}, \;
    \appliedname{Tianshuo Xu\textsuperscript{1}}, \;
    \appliedname{Narendra Ahuja\textsuperscript{2}}, \;
    \appliedname{Wei Zhan\textsuperscript{1,4,\dag}}\par}
   \smallskip
   {\small\rmfamily\color{AppliedGray}%
    \textsuperscript{1}Applied Intuition \quad
    \textsuperscript{2}University of Illinois Urbana-Champaign\\
    \textsuperscript{3}Purdue University \quad
    \textsuperscript{4}University of California, Berkeley\par}
   \smallskip
   {\footnotesize\rmfamily\color{AppliedGray}%
    \textsuperscript{*}Work done during an internship at Applied Intuition \quad
    \textsuperscript{\dag}Corresponding author: \href{mailto:wei.zhan@applied.co}{wei.zhan@applied.co}\par}
   \smallskip
   {\small\rmfamily
    \href{https://s2tok.github.io/}{\textbf{Project page:} \nolinkurl{https://s2tok.github.io/}}\par}
   \par}
}
\begin{abstract}

Streaming 3D reconstruction requires more than a sequence of geometric predictions: it requires a persistent scene state that can incorporate new evidence and remain renderable as observations arrive. Latent spatial tokens offer a promising representation for this purpose, but constructing them from an image collection leaves open how to maintain them online, where each observation may both revisit known regions and reveal new content. We introduce S2Tok, a feed-forward framework that maintains a size-adaptive, persistent scene state from uncalibrated image streams. Its central idea is to distinguish updates to the existing representation from selective expansion. A spatially informed transformer integrates each incoming observation with the persistent scene tokens, while a learned admission module selectively expands the representation to limit redundant storage. A hierarchical decoder and Gaussian head convert the evolving state into non-pixel-aligned 3D Gaussians, enabling novel-view rendering without caching previous frames. Experiments across four benchmarks demonstrate competitive streaming rendering quality with compact Gaussian representations. These results support latent spatial tokens as a persistent computational state for online 3D reconstruction, combining learned scene updates with explicit Gaussian rendering.
\end{abstract}
\makefrontmatter
\section{Introduction}
\label{intro}
Streaming 3D reconstruction aims to recover a scene as images arrive, rather than after
a complete video sequence. For interactive scene capture and
embodied mapping, a reconstruction is useful before observation is complete: it can help
assess scene coverage and inform the next viewpoint or action. Each new observation may
reveal unseen structure, resolve earlier ambiguities, or provide better evidence about
previously observed regions. A useful reconstruction system must therefore maintain a
representation of the accumulated scene that can be revised online and rendered
throughout the process. This makes it a problem of maintaining scene state,
not merely producing successive geometric predictions. Latent spatial tokens are
appealing because they provide a shared feature space for integrating evidence across
views, separating the learned scene memory from the explicit geometry used for
rendering. We investigate how to maintain and update such a tokenized scene state online
while keeping it decodable into an explicit 3D scene. 

Existing work offers two complementary capabilities. Token-based methods such as
ZipSplat~\citep{zipsplat}, TokenGS~\citep{tokengs}, and GlobalSplat~\citep{globalsplat}
decode latent scene tokens into non-pixel-aligned Gaussians, but construct these tokens
by jointly processing a provided set of images. Recurrent methods such as
CUT3R~\citep{cut3r} and Point3R~\citep{point3r} instead update persistent memory as
images arrive. Their standard streaming pipelines, however, use this memory to predict
pointmaps for each view and form the scene by accumulating those predictions.
We seek to connect these capabilities through a size-adaptive scene-token state that is updated online and jointly decoded into the accumulated Gaussian scene at each step. This places recurrent scene maintenance and explicit scene rendering within a shared representation.

Maintaining an adaptive scene state requires separating state updates from expansion. A new view of an already observed surface may provide additional evidence about its geometry or appearance without introducing a new region, whereas a change in viewpoint may reveal previously occluded content. The former can be handled by updating existing tokens; the latter may require additional representational capacity. Retaining all incoming tokens conflates these roles and accumulates redundant information across overlapping views, while restricting the state to a fixed token set can create a capacity bottleneck as more of the scene is observed. Moreover, without known camera parameters, the model must infer how incoming observations correspond to the accumulated scene. The desired representation should therefore support both spatially informed updates to existing tokens and selective admission of additional tokens, rather than equating every new observation with an expansion of scene memory.

We introduce \textbf{S2Tok}, a feed-forward framework for streaming 3D Gaussian reconstruction of static scenes from uncalibrated image streams. S2Tok maintains an adaptive set of latent spatial tokens as the persistent scene state and separates updates to existing tokens from selective expansion. A Stream-Spatial Transformer integrates incoming frame tokens into the existing spatial token space and updates the spatial tokens according to the new frame. A learned admission module then selects the frame tokens predicted to carry novel content. A hierarchical decoder and Gaussian head map the resulting tokens to non-pixel-aligned Gaussians in a shared reference frame. Because the accumulated scene is decoded anew at each step, incoming evidence can update the representation of previously observed regions rather than only contribute additional framewise outputs. The updated spatial tokens are carried forward without caching previous frames, supporting continued reconstruction and novel-view rendering.


A geometric novelty criterion provides useful guidance for token admission, but does not
directly measure how retaining a token affects the reconstructed scene. Because
admission decisions are discrete, rendering losses cannot directly train the policy
through ordinary backpropagation. We therefore bootstrap the admission module by
distilling a depth-based geometric novelty oracle, then refine it using self-critical
sequence training. This refinement compares the reconstruction quality under sampled
admissions with that under the policy's deterministic decisions, providing
reconstruction-driven feedback for discrete selection. The policy can thus adjust its
admission decisions according to their effect on reconstruction, rather than remaining
solely tied to geometric novelty labels.


Our contributions are threefold:
\begin{itemize}[topsep=0pt, leftmargin=12pt]
    \setlength{\itemsep}{2pt}
    \setlength{\parskip}{2.5pt}
    \item \textbf{Persistent scene-state representation.}
    We introduce a feed-forward streaming framework that maintains a size-adaptive set of latent spatial tokens and jointly decodes the accumulated Gaussian scene from the current state, without retaining past images at inference time.

    \item \textbf{Decoupled state updating and expansion.}
    We separate observation integration from state growth through spatially informed recurrent updates and a learned token-admission policy. The policy is initialized with geometric supervision and subsequently optimized using reconstruction-driven policy gradients.

    \item \textbf{Compact streaming reconstruction.}
    Across four benchmarks, we evaluate both intermediate and final reconstructions, demonstrating competitive rendering performance with compact Gaussian representations.
\end{itemize}

\insertpaperteaser
\section{Related Work}
\label{sec:related}
\paragraph{Offline 3D Reconstruction.}
Recent advances in 3D reconstruction cover a broad range of static and dynamic settings~\citep{dust3r,sc4dgs,limr,ros-cam,s3o,vggt,helix4d}.
Among feed-forward approaches, DUSt3R~\citep{dust3r} formulates uncalibrated reconstruction as pairwise pointmap regression, with global alignment for larger image collections, while VGGT~\citep{vggt} jointly predicts camera parameters and dense geometry from multiple views.
For renderable scene reconstruction, pixelSplat~\citep{pixelsplat}, MVSplat~\citep{mvsplat}, and DepthSplat~\citep{depthsplat} predict pixel-aligned Gaussian primitives from posed images.
NoPoSplat~\citep{noposplat} removes the requirement for input camera poses by predicting Gaussians directly in a shared canonical coordinate system. UniQueR~\citep{uniquer} moves away from pixel-aligned Gaussian prediction by using 3D queries that decode into Gaussians. These methods jointly process images without a persistent scene representation that evolves with an incoming stream. Most of those relying on pixel-aligned predictions also incur higher storage costs.

\paragraph{Streaming 3D Reconstruction.}
Optimization-based monocular Gaussian SLAM methods~\citep{matsuki2024gaussian} continuously estimate camera poses and update explicit Gaussian maps from RGB streams, but rely on iterative test-time tracking and map optimization.
Recent feed-forward methods instead integrate observations through causal attention or recurrent state updates.
StreamVGGT~\citep{streamvggt} caches historical keys and values for causal geometric inference, while IncVGGT~\citep{incvggt} compresses and prunes historical representations to bound memory consumption.
Recurrent alternatives, including CUT3R~\citep{cut3r}, Point3R~\citep{point3r}, and Point4Cast~\citep{point4cast}, integrate observations into persistent latent or spatially anchored states and produce dense geometric predictions.
For renderable streaming reconstruction, StreamGS~\citep{streamgs} predicts and aggregates per-frame Gaussians, using adjacent-frame correspondences for feature refinement and redundant Gaussian merging.
Flash-Mono~\citep{flashmono} recurrently aggregates visual features to predict camera poses and Gaussian surfels for monocular SLAM, while StreamSplat~\citep{streamsplat} performs feed-forward Gaussian prediction and adaptive fusion over uncalibrated video streams.
Our focus is on maintaining a size-adaptive latent scene state that is updated with incoming observations and directly decoded into non-pixel-aligned Gaussians, without retaining a historical KV cache.

\paragraph{Latent Spatial Tokens for 3D Reconstruction.}
Latent spatial representations offer an alternative to tying scene primitives to input pixels.
In object-centric reconstruction, LRM~\citep{lrm} uses a transformer to lift image features into structured triplane features for radiance-field reconstruction.
More directly related to Gaussian scene reconstruction, TokenGS~\citep{tokengs} introduces learnable scene tokens and directly regresses 3DGS coordinates, decoupling primitive count from input resolution and the number of views.
GlobalSplat~\citep{globalsplat} aggregates multi-view information into a fixed set of global scene tokens before decoding explicit Gaussians.
These token-based Gaussian models reduce redundancy through scene-level aggregation, but their feed-forward reconstruction pipelines operate on a supplied observation set rather than incrementally maintaining and expanding a persistent scene-token memory.
S2Tok instead maintains these tokens recurrently, combining observation-conditioned updates with selective admission of additional tokens. This brings latent-token Gaussian decoding into a pose-free streaming framework that separates scene-state updating from representation growth.
\section{Method}
\label{sec:methods}

\begin{figure}[t]
    \centering
    \includegraphics[width=\linewidth,trim={17.92bp 8.29bp 25.23bp 3.97bp},clip]{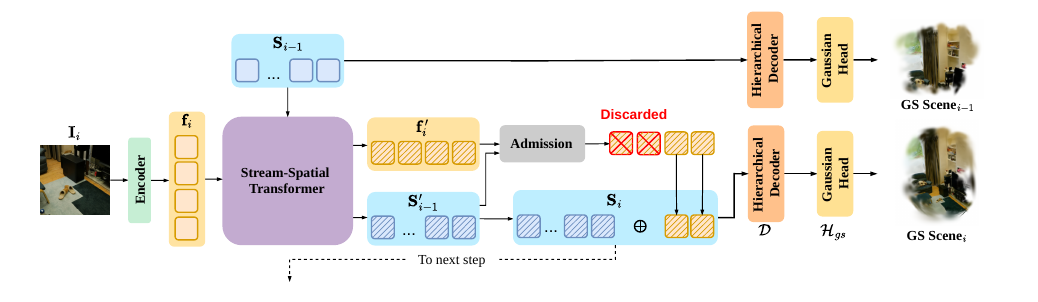}
    \caption{\textbf{S2Tok Pipeline.} The encoder maps the new frame $\mathbf{I}_i$ to tokens $\mathbf{f}_i$. The SS-Transformer jointly updates frame tokens and existing spatial tokens $\mathbf{S}_{i-1}$. The admission module appends selected frame tokens to form $\mathbf{S}_i$, which is carried to the next step. The Hierarchical Decoder and Gaussian Head decode and predict Gaussian attributes. $\oplus$ denotes concatenation.}
    \label{fig:pipeline}
\end{figure}


We process a stream of images sequentially. At step $i$, the incoming image $\mathbf{I}_i \in \mathbb{R}^{H \times W \times 3}$ is used to update the latent scene state and produce the current accumulated Gaussian reconstruction.
As shown in Fig. \ref{fig:pipeline}, we first encode $\mathbf{I}_i$ into frame tokens $\mathbf{f}_i$\footnote{Such frame tokens are the representations implicitly encoded in their own frame's camera coordinate system.}. Secondly, we feed the existing latent spatial tokens $\mathbf{S}_{i-1}$ from the last step and the newly encoded $\mathbf{f}_i$ into the proposed Stream-Spatial (SS) Transformer. In the SS-Transformer, $\mathbf{S}_{i-1}$ is adapted to the new frames and viewpoints; meanwhile, $\mathbf{f}_i$ is implicitly aligned to the latent spatial space.
We then apply the admission module to the updated $\mathbf{f}'_i$ and $\mathbf{S}'_{i-1}$, and append the selected frame tokens to form $\mathbf{S}_{i} \in \mathbb{R}^{(M+\delta) \times D}$,
where $\delta$ is the number of admitted frame tokens. We then provide $\mathbf{S}_{i}$ and token sets from two other layers of the SS-Transformer as inputs to the Hierarchical Decoder and predict Gaussian candidates with the Gaussian Head. Each token corresponds to $h$ Gaussians. Finally, $\mathbf{S}_{i}$ is carried over to the next step.

\subsection{Encoder}
\label{sec:encoder}
At each step (Fig.~\ref{fig:pipeline}), a ViT encoder~\citep{vit} patchifies and encodes $\mathbf{I}_i$ as $\mathbf{f}_i = \mathrm{Encoder}(\mathbf{I}_i) \in \mathbb{R}^{\frac{H}{w}\times\frac{W}{w}\times D}$, where $w$ is the patch size and $D$ is the feature dimension.


\subsection{Stream-Spatial Transformer}
\label{sec:ss-transformer}
\begin{figure}[tbp]
    \centering
    \begin{minipage}{0.60\linewidth}
    \centering
    \includegraphics[width=\linewidth]{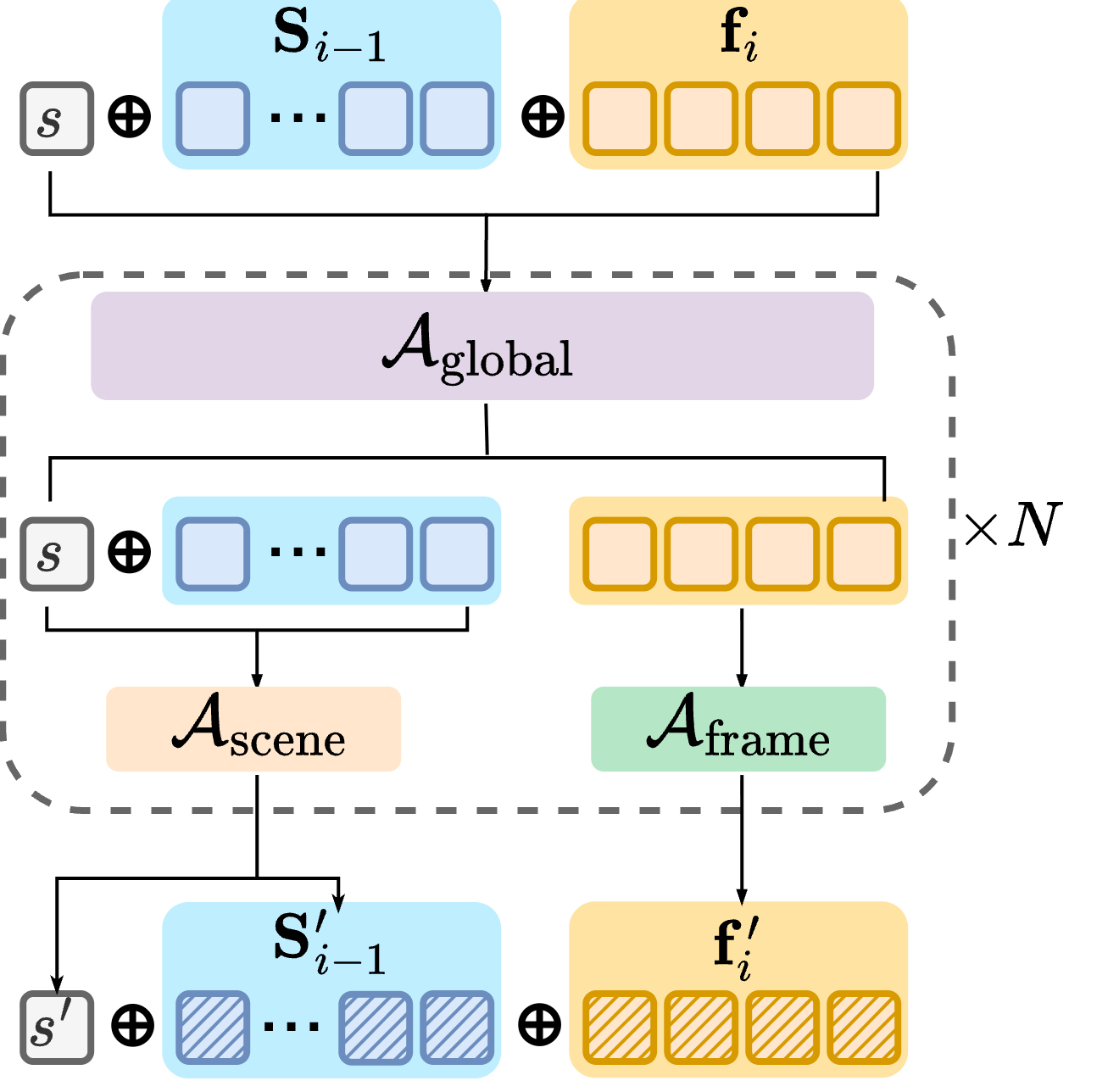}
    \caption{\textbf{Stream-Spatial Transformer.} Scale token $s$, spatial tokens $\mathbf{S}_{i-1}$, and frame tokens $\mathbf{f}_i$ attend to one another in $\mathcal{A}_{\text{global}}$. Scene-local attention $\mathcal{A}_{\text{scene}}$ then updates $s$ and $\mathbf{S}_{i-1}$, while frame-local attention $\mathcal{A}_{\text{frame}}$ updates $\mathbf{f}_i$.}
    \label{fig:ss-transformer}
\end{minipage}
\end{figure}
The existing latent spatial tokens $\mathbf{S}_{i-1} \in \mathbb{R}^{M \times D}$, where $M$ is the number of spatial tokens, contain scene information accumulated from observations up to step $i-1$. (The accumulation process is discussed in Sec.~\ref{sec:admission}.) However, $\mathbf{S}_{i-1}$ and $\mathbf{f}_i$ are implicitly encoded in the shared reference and local camera coordinates, respectively.
Thus, we propose the Stream-Spatial Transformer to adapt $\mathbf{S}_{i-1}$ to the new observation, and implicitly align $\mathbf{f}_i$ to the reference coordinate of $\mathbf{S}_{i-1}$.
As depicted in Fig.~\ref{fig:ss-transformer}, SS-Transformer has $N$ blocks, each consisting of a global attention layer $\mathcal{A}_{\text{global}}$, a scene-local attention layer $\mathcal{A}_{\text{scene}}$, and a frame-local attention layer $\mathcal{A}_{\text{frame}}$. These three layers share the same architecture but operate under different settings. Before the SS-Transformer, we concatenate a learnable normalization scale token $s$ with $\mathbf{S}_{i-1}$.

\paragraph{Normalization Scale Token.} 
At each step, S$^{2}$Tok directly predicts Gaussian candidates\footnote{Each Gaussian candidate contains a mean $\mu\in\mathbb{R}^{3}$, scale $s^{gs}\in\mathbb{R}^{3}$, quaternion $q\in\mathbb{R}^{4}$, opacity $o\in\mathbb{R}$, and SH coefficients $c^{sh}\in\mathbb{R}^{4\times 3}$ (degree 1), i.e., 23 parameters in total.} in the same reference coordinate (the coordinate of $\mathbf{I}_0$ by default). However, the spatial extent of the observed scene grows as new frames arrive. To mitigate scale-related out-of-distribution effects, predictions at each step should all be expressed in the normalized space, requiring step-dependent normalization scales. We therefore introduce a learnable normalization scale token $s$ to help the model implicitly account for variations in scene scale across  steps.
Appendix~\ref{sec:scale_gt} describes how we generate pseudo-GT scales and use them to normalize supervision at each step.
The ablation results in Tab.~\ref{tab:dl3dv-inference-ablation} demonstrate its benefits for model learning.

\paragraph{Global Attention Layer.} We leverage a global attention layer $\mathcal{A}_{\text{global}}$ to let $s$ attend to $\mathbf{S}_{i-1}$ and $\mathbf{f}_i$ to be aware of the relations between the existing scene and the new frame, adapt $\mathbf{S}_{i-1}$ to the new observation, and align $\mathbf{f}_i$ into the latent spatial space. Specifically, in  $\mathcal{A}_{\text{global}}$, we do not add positional encoding. All tokens passed into $\mathcal{A}_{\text{global}}$ attend to one another as follows:
\begin{equation}
\mathcal{A}_{\text{global}}(s \oplus \mathbf{S}_{i-1} \oplus \mathbf{f}_i)
\end{equation}

\paragraph{Scene-local Attention Layer.} To let $\mathbf{S}_{i-1}$ capture the spatial relations within the existing scene, we adopt a scene-local attention layer $\mathcal{A}_{\text{scene}}$ for existing scene updates only.
$\mathcal{A}_{\text{scene}}$ takes only $\mathbf{S}_{i-1}$ and $s$ as input.
To encode spatial relationships existing in the latent spatial tokens $\mathbf{S}_{i-1}$, we apply a 3D variant of rotary position embedding (RoPE)~\citep{rope} to $\mathbf{S}_{i-1}$:
\begin{equation}
\mathcal{A}_{\text{scene}}(s \oplus \mathrm{RoPE_{3D}}(\mathbf{S}_{i-1}))
\end{equation}
We use the opacity-weighted geometric median of Gaussian centers of each spatial token predicted at step $i-1$ as the 3D positional coordinates for $\mathrm{RoPE_{3D}}$. The 
geometric-median calculation is detailed in Appendix~\ref{sec:weightedaverage}.

\paragraph{Frame-local Attention Layer.} 
The frame-local attention layer updates only $\mathbf{f}_i$, using 2D RoPE with the tokens' patch-grid coordinates as positional inputs:
\begin{equation}
\mathcal{A}_{\text{frame}}(\mathrm{RoPE_{2D}}(\mathbf{f}_{i}))
\end{equation}
Overall, we represent the outputs of the Stream-Spatial Transformer as $s'$, $\mathbf{S}'_{i-1}$, and $\mathbf{f}'_i$.

\subsection{Learned Admission for Selective State Expansion}
\label{sec:admission}
\paragraph{Learnable Admission Module.} To control the growth of the scene-token state, we introduce a learnable admission module consisting of a cross-attention layer and a self-attention layer. At each step, it takes $\mathbf{f}'_i$ as query, $\mathbf{S}'_{i-1}$ as key/value, and predicts admission logits $\pi$. Frame tokens whose admission probabilities exceed $\tau$ are appended to the updated scene tokens:
\begin{equation}
    \mathbf{S}_i = \mathbf{S}'_{i-1} \oplus \mathbf{f}'_{i}[\sigma(\pi) > \tau], \quad \pi = \mathrm{Admission}(query=\mathbf{f}'_i, key/value=\mathbf{S}'_{i-1})
\end{equation}
Here, $\sigma$ is the Sigmoid function. The resulting $\mathbf{S}_i$ is carried to step $i+1$, allowing $\mathbf{f}_{i+1}$ to interact with the accumulated scene representation without storing a historical KV cache.

\paragraph{Two-stage Learning Strategy.}
At the start of training, directly using the admission module’s outputs for hard token selection prevents ordinary backpropagation through the admission decisions to the admission logits, potentially reinforcing arbitrary early decisions. Such discrete selections prevent gradients from reaching the admission logits, which can push arbitrary early decisions and lead to a poor local optimum. We therefore train the admission module in
two stages. First, we distill the decisions of a depth-based geometric oracle (Appendix~\ref{sec:admitpseudogt}) using the class-weighted binary cross-entropy loss discussed in Sec.~\ref{sec:loss}. 

Oracle supervision provides an initialization for admission, but does not optimize the policy according to reconstruction outcomes. We therefore further optimize the admission policy using self-critical sequence training (SCST)~\citep{scst}, which compares sampled and deterministic admissions through their reconstruction errors. This provides reconstruction-driven policy gradients for discrete admission decisions. 


Specifically, SCST performs two feed-forward rollouts under the same
model parameters. Let
$\mathbf{p}=\sigma(\pi)$ denote the predicted admission
probabilities. The first rollout serves as a self-critical baseline
without gradient computation and applies the deterministic admission
mask $\mathbf{a}^{d}$, whereas the second rollout applies a stochastic
admission mask $\mathbf{a}^{b}$ sampled from the Bernoulli
policy:
\begin{equation}
    \mathbf{a}^{d}
    =
    \mathbbm{1}
    \left[
        \mathbf{p}>\tau
    \right],
    \qquad
    \mathbf{a}^{b}
    \sim
    \operatorname{Bernoulli}(\mathbf{p}).
\end{equation}

Both $\mathbf{a}^{d}$ and $\mathbf{a}^{b}$ are binary masks that can
admit multiple tokens. We denote the reconstruction errors produced by
the two rollouts as $e^{d}$ and $e^{b}$, respectively, and define the
self-critical advantage as $A=e^{b}-e^{d}$. Since these quantities
measure reconstruction error, $A<0$ indicates that the sampled
admission decisions outperform the deterministic baseline. The
admission module is then optimized using the SCST objective discussed
in Sec.~\ref{sec:loss}. In practice, both the advantage and the log-probability are computed per step with a reward-to-go formulation; the step-indexed form of Eq.~\ref{eq:scst} and all stabilization details are discussed in
Appendix~\ref{sec:training-details}.

\begin{figure}[tbp]
    \centering
    \begin{minipage}{0.60\linewidth}
    \centering
    \includegraphics[
        width=\linewidth,
        height=0.75\linewidth,
        keepaspectratio,
        trim={18.24bp 15.27bp 21.16bp 15.22bp},
        clip
    ]{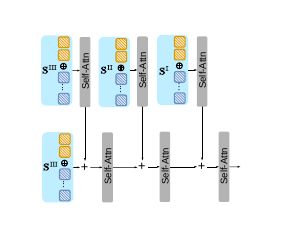}

    \caption{\textbf{Hierarchical Decoder.} Hierarchical tokens $\mathbf{S}^\mathrm{III}$, $\mathbf{S}^\mathrm{II}$, and $\mathbf{S}^\mathrm{I}$ from the SS-Transformer are processed individually through self-attention layers and then fused through residual addition \textbf{+} into the decoding stream, which is initialized from $\mathbf{S}^{\mathrm{III}}$. A self-attention layer allows one more update after each residual addition.}
    \label{fig:multi-decoder-head}
\end{minipage}
\end{figure}
\subsection{Hierarchical Decoder \& Head}
\label{sec:mult-decoder-head}
We update the hierarchical decoder~\citep{zipsplat} with self-attention only to jointly decode the current accumulated scene from the updated token representation, as shown in Fig.~\ref{fig:multi-decoder-head}.
Our Hierarchical Decoder $\mathcal{D}$ takes as input $\mathbf{S}^{\mathrm{I}}$, $\mathbf{S}^{\mathrm{II}}$, and $\mathbf{S}^{\mathrm{III}}$(${=}\,\mathbf{S}_{i}$), three layers' token sets from SS-Transformer, each formed by concatenating the spatial tokens with the admitted ($\sigma(\pi)>\tau$) frame tokens at the same layer, and predicts Gaussians $\mathbf{G} \in \mathbb{R}^{h(M+\delta) \times 23}$ using the Gaussian Head $\mathcal{H}_{gs}$:
\begin{equation}
    \mathbf{G} = \mathcal{H}_{gs}(\mathcal{D}(\mathbf{S}^{\mathrm{I}}, \mathbf{S}^{\mathrm{II}}, \mathbf{S}^{\mathrm{III}}))
\end{equation}

\subsection{Losses}
\label{sec:loss}
\paragraph{Rendering Loss.}
At each step, we render the accumulated scene in the normalized space. Specifically, the pseudo-GT scale $\ell$ is derived from the median distance of the accumulated point cloud to the reference coordinate origin. Rendering supervision is applied on both the newly and previously observed views.
Each supervision view contributes an RGB L1 loss $\mathcal{L}_{rgb}$, a perceptual loss $\mathcal{L}_{lpips}$, and a depth L1 loss $\mathcal{L}_{depth}$. The per-step rendering loss is
\begin{equation}
\mathcal{L}_{render}= \mathcal{L}_{rgb} + \lambda_{lpips}\mathcal{L}_{lpips} + \lambda_{depth}\mathcal{L}_{depth} +\mathcal{R},
\end{equation}
where $\mathcal{R}$ is a regularization term on the predicted $\mathbf{G}$: penalties on scale $>0.1$, opacity $<0.01$, and centers outside $[-10,10]^3$. In addition, at every step we also decode $\mathbf{f}'_{i}$ alone, and supervise this solo rendering on $\mathbf{I}_i$ with the same recipe $\lambda^{solo}_{render}\mathcal{L}_{render}^{solo}$ to preserve the per-frame gradients.

\paragraph{Chamfer Loss.}
The image-space reconstruction terms do not directly supervise the set $\mathbf{G}^{nr}$ of Gaussians never rendered by any supervised view. We therefore adopt a single-direction Chamfer loss $\mathcal{L}_{chamfer}$ that only pulls never-rendered Gaussian centers $\mu^{nr}$ toward the point cloud $\mathbf{P}$ unprojected from pseudo-GT camera and depth. The solo pass carries its own $\lambda^{solo}_{chamfer}\mathcal{L}_{chamfer}^{solo}$ over all Gaussians. We have:
\begin{equation}
\mathcal{L}_{chamfer}
=\frac{1}{|\mathbf{G}^{nr}|}\sum_{k\in\mathbf{G}^{nr}}
\min_{p\in \mathbf{P}}\big\lVert \mu^{nr}_k - p \big\rVert_2
\end{equation}

\paragraph{Normalization Scale Loss.}
An MLP head predicts the log normalization scale $\hat{\ell}$ from $s'$ from SS-Transformer. The predictions are then supervised by the normalization scale loss $\mathcal{L}_{scale}$ ($\ell$ is the pseudo-GT scale calculated in Sec.~\ref{sec:scale_gt}): $\mathcal{L}_{scale}=\lambda_{scale}\big(\hat{\ell}-\log \ell\big)^2$.

\paragraph{Class-weighted Binary Cross-entropy Loss.}
For all frame tokens in $\mathbf{f}'_i$, the admission head produces logits $\pi$, which are then supervised by a class-weighted binary cross-entropy $\mathcal{L}_{bce}$ against pseudo-GT admission labels
$y \in \{0,1\}$:
\begin{equation}
\mathcal{L}_{bce}
=-\frac{w^2}{HW}\sum_{j=1}^{HW/w^2}
\Big[\beta\,y_j\log\sigma(\pi_j)
+\big(1-y_j\big)\log\big(1-\sigma(\pi_j)\big)\Big],
\end{equation}
where $\beta$ is an empirically decided positive-class weight. The gradient flows through the head's inputs
into the backbone, letting the representation co-adapt to the admission
task; at inference the head admits patches with $\sigma(\pi)> \tau$.

\paragraph{SCST Loss.}
SCST optimizes the admission module using the policy gradient. Let $P_{\theta}(\mathbf{a}^b)$ denote the probability of producing the
sampled binary admission mask $\mathbf{a}^b$. We define its normalized log-probability as $\Theta(\mathbf{a}^b)=\frac{1}{|\mathbf{a}^b|}\log P_{\theta}(\mathbf{a}^b)$, where $|\mathbf{a}^b|$ is the number of binary admission decisions in the sampled rollout. The objective is thus a policy-gradient loss whose
gradient is weighted by reconstruction performance:
\begin{equation}
\label{eq:scst}
    \mathcal{L}_{scst}
    =
    \frac{\operatorname{sg}(A)}
         {\hat{\sigma}_{A}}
    \Theta(\mathbf{a}^b) + \mathcal{L}_{KL},
\end{equation}
where $\operatorname{sg}(\cdot)$ denotes stop-gradient and
$\hat{\sigma}_{A}$ is a running estimate of the advantage standard deviation. To stabilize early policy optimization while allowing the learned policy to depart from the geometric oracle,
we also introduce KL-divergence regularization $\mathcal{L}_{KL} = \beta_{KL} D_{\mathrm{KL}}\left(Q\,\Vert\,P_{\theta}\right)$, where $Q$ denotes the smoothed oracle admission distribution and $P_{\theta}$ denotes the Bernoulli distribution parameterized by the admission probabilities predicted by the admission head. 

\section{Experiments}
\label{sec:experiments}
We train S2Tok solely on DL3DV~\citep{dl3dv} and RealEstate10K~\citep{rel10k}. More training, testing, and implementation details can be found in Sec.~\ref{sec:implementation} in the Appendix. 
For evaluations, we show extensive quantitative and qualitative comparisons between S2Tok and other baselines.
We get the rendering performance of pointmap prediction methods using point-splat rendering~\citep{zwicker2001surface, pfister2000surfels}.

Specifically, we hold out one target view between consecutive context
views and evaluate reconstruction quality in two settings:
1) \textbf{Final}: Render all target views using the final
    reconstruction after processing the whole context sequence.
2) \textbf{Intermediate}: After each step, render the
    corresponding target view using the current reconstruction,
    based only on the context views observed so far.
Streaming methods update the scene incrementally, whereas non-streaming
methods jointly process all available context views anew at each
evaluation step. StreamSplat is evaluated only under \textbf{Intermediate}, as its
time-conditioned reconstruction does not produce a globally
camera-aligned final scene.

\begin{table*}[htbp]
    \centering
    \caption{Quantitative Comparison on DL3DV. Each metric cell reports PSNR$\uparrow$ / SSIM$\uparrow$ / LPIPS$\downarrow$. $r{=}1$ denotes the default compression ratio of ZipSplat. $^{\dagger}$ adjusts the ratio to match our token budget. \textbf{Avg.} represents the average across all sequences. \textbf{Time} and \textbf{Mem.} report per-view reconstruction time and peak GPU memory of the streaming methods in the 50-view setting, measured on a single NVIDIA A100 80GB. Notably, MonoGS is not feed-forward.}
    \label{tab:dl3dv_comparison}

    \renewcommand{\arraystretch}{1.05}
    \resizebox{\textwidth}{!}{%
    \begin{tabular}{
        @{}l
        @{\hspace{4pt}}c
        @{\hspace{4pt}}c
        @{\hspace{8pt}}ccc
        @{\hspace{8pt}}cc
        @{\hspace{8pt}}c
        @{\hspace{4pt}}c@{}
    }
        \toprule[1.3pt]

        \multirow{3}{*}{\textbf{Method}}
        & \multirow{3}{*}{\shortstack{\textbf{Stream-}\\\textbf{ing}}}
        & \multirow{3}{*}{\shortstack{\textbf{Cam-}\\\textbf{free}}}
        & \multicolumn{3}{c}{\textbf{Final}}
        & \multicolumn{2}{c}{\textbf{Intermediate}}
        & \multirow{3}{*}{\shortstack{\textbf{Time}\\\textbf{(ms/v)}}}
        & \multirow{3}{*}{\shortstack{\textbf{Mem.}\\\textbf{(GiB)}}} \\
        \cmidrule(lr){4-6}
        \cmidrule(lr){7-8}

        & & &
        \multirow{2}{*}{\textbf{12 views}}
        & \multirow{2}{*}{\textbf{50 views}}
        & \textbf{Avg. \#GS/PT}
        & \multirow{2}{*}{\textbf{12 views}}
        & \multirow{2}{*}{\textbf{50 views}}
        & & \\

        & & & & &
        \textbf{12 / 50 views}
        & & & & \\

        \midrule

        ZipSplat$^{r{=}1}$
        & \xmark & \cmark
        & 23.04/0.69/0.24
        & 21.94/0.63/0.28
        & 124K/518K
        & 23.26/0.70/0.25
        & 22.13/0.62/0.29
        & -- & -- \\

        ZipSplat$^{\dagger}$
        & \xmark & \cmark
        & 21.79/0.64/0.31
        & 20.88/0.58/0.35
        & 36K/66K
        & 21.74/0.64/0.32
        & 20.81/0.57/0.37
        & -- & -- \\

        TokenGS
        & \xmark & \xmark
        & 21.74/0.69/0.40
        & 20.53/0.63/0.50
        & 262K/262K
        & 22.64/0.73/0.34
        & 21.54/0.65/0.45
        & -- & -- \\

        GlobalSplat
        & \xmark & \xmark
        & 17.71/0.47/0.55
        & 15.92/0.40/0.63
        & 32K/32K
        & 18.65/0.52/0.49
        & 17.01/0.44/0.58
        & -- & -- \\

        AnySplat
        & \xmark & \cmark
        & 18.65/0.54/0.30
        & 19.35/0.55/0.30
        & 686K/2.6M
        & 17.93/0.53/0.32
        & 19.47/0.57/0.29
        & -- & -- \\

        DepthSplat-B
        & \xmark & \xmark
        & 17.84/0.56/0.37
        & 15.52/0.44/0.47
        & 762K/3.2M
        & 20.46/0.66/0.27
        & 18.50/0.56/0.35
        & -- & -- \\

        \midrule

        MonoGS
        & \cmark & \cmark
        & 13.30/0.29/0.64
        & 13.83/0.32/0.69
        & 8K/13K
        & 14.42/0.33/0.59
        & 13.80/0.31/0.66
        & 2255 & 1.3 \\

        \midrule


        CUT3R
        & \cmark & \cmark
        & 12.52/0.19/0.68
        & 12.07/0.11/0.74
        & 762K/3.2M
        & 11.90/0.18/0.68
        & 10.38/0.11/0.74
        & 41 & 4.8 \\

        StreamSplat
        & \cmark & \cmark
        & --
        & --
        & 750K/3.2M
        & 16.77/0.37/0.48
        & 17.93/0.40/0.45
        & 225 & 5.7 \\


        Spann3R
        & \cmark & \cmark
        & 13.08/0.16/0.66
        & 12.22/0.09/0.71
        & 762K/3.2M
        & 12.04/0.12/0.68	
        & 11.96/0.11/0.71
        & 84 & 2.8 \\


        Point3R
        & \cmark & \cmark
        & 12.03/0.13/0.69
        & 10.62/0.07/0.73
        & 762K/3.2M
        & 11.52/0.12/0.70
        & 10.14/0.07/0.71
        & 260 & 4.7 \\


        StreamVGGT
        & \cmark & \cmark
        & 11.10/0.12/0.74
        & 10.10/0.04/0.79
        & 762K/3.2M
        & 9.78/0.09/0.74
        & 10.10/0.05/0.80	
        & 197 & 13.1 \\

        \textbf{S2Tok (Ours)}
        & \cmark & \cmark
        & \textbf{19.52/0.52/0.43}
        & \textbf{18.00/0.42/0.51}
        & \textbf{36K/66K}
        & \textbf{20.27/0.55/0.40}
        & \textbf{19.11/0.46/0.47}
        & 81 & 7.8 \\

        \bottomrule[1.3pt]
    \end{tabular}%
    }
\end{table*}

\paragraph{DL3DV.} 
The evaluation sequences include substantial camera motion. In Tab.~\ref{tab:dl3dv_comparison}, S2Tok achieves the highest PSNR among the compared feed-forward streaming methods in both the Final and Intermediate settings, using 36K and 66K Gaussians for the 12-view and 50-view evaluations, respectively. Its advantages are not uniform across all rendering metrics. Fig.~\ref{fig:dl3dv_final} and Fig.~\ref{fig:dl3dv_causal} illustrate the resulting reconstructions, including examples with fewer visible foreground–background artifacts than several compared baselines.


\paragraph{RealEstate10K.} Tab.~\ref{tab:re10k_mipnerf360_comparison} and Tab.~\ref{tab:re10k_comparison-full} show that S2Tok achieves the highest PSNR among the compared feed-forward streaming methods in both evaluation settings, while using 17K and 21K Gaussians for the 12-view and 50-view evaluations, respectively. These results demonstrate compact scene reconstruction on the evaluated indoor sequences.
\input{figures/fig_qual_dl3dv_final.tex}

\begin{table*}[htbp]
    \centering
    \caption{Quantitative Comparison on RealEstate10K and Mip-NeRF 360. Each metric cell reports PSNR$\uparrow$ / SSIM$\uparrow$ / LPIPS$\downarrow$. Results are evaluated under 12-view and 50-view settings. Detailed comparisons are in Tab.~\ref{tab:re10k_comparison-full} and Tab.~\ref{tab:mipnerf360_comparison-full} in the Appendix. $r{=}1$ is the default compression ratio of ZipSplat. $^{\dagger}$ adjusts the compression ratio to match our token budget. Notably, MonoGS is not feed-forward.}
    \label{tab:re10k_mipnerf360_comparison}

    \renewcommand{\arraystretch}{1.08}
    \setlength{\tabcolsep}{4pt}

    \resizebox{\textwidth}{!}{%
    \begin{tabular}{
        @{}l
        c
        c
        c
        c
        c
        c
        c
        c@{}
    }
        \toprule[1.3pt]

        \multirow{3}{*}{\textbf{Method}}
        & \multirow{3}{*}{\shortstack[c]{\textbf{Stream-}\\\textbf{ing}}}
        & \multirow{3}{*}{\shortstack[c]{\textbf{Cam-}\\\textbf{free}}}
        & \multicolumn{3}{c}{\textbf{RealEstate10K (12 Views)}}
        & \multicolumn{3}{c}{\textbf{Mip-NeRF 360 (50 Views)}} \\

        \cmidrule(lr){4-6}
        \cmidrule(lr){7-9}

        & & &
        \textbf{Final}
        & \shortstack[c]{\textbf{Avg.}\\\textbf{\#GS/PT}}
        & \textbf{Intermediate}
        &
        \textbf{Final}
        & \shortstack[c]{\textbf{Avg.}\\\textbf{\#GS/PT}}
        & \textbf{Intermediate} \\

        \midrule

        ZipSplat$^{r{=}1}$
        & \xmark
        & \cmark
        & 27.65/0.87/0.13
        & 124K
        & 27.95/0.88/0.12
        & 21.46/0.54/0.38
        & 518K
        & 21.39/0.53/0.38 \\

        ZipSplat$^{\dagger}$
        & \xmark
        & \cmark
        & 25.47/0.82/0.19
        & 17K
        & 26.26/0.84/0.17
        & 20.66/0.50/0.43
        & 82K
        & 20.68/0.49/0.43 \\

        TokenGS
        & \xmark
        & \xmark
        & 32.00/0.94/0.15
        & 262K
        & 30.79/0.90/0.17
        & 22.29/0.64/0.48
        & 262K
        & 22.88/0.67/0.43 \\

        GlobalSplat
        & \xmark
        & \xmark
        & 29.85/0.91/0.13
        & 32K
        & 29.23/0.89/0.14
        & 17.25/0.39/0.65
        & 32K
        & 17.85/0.42/0.59 \\

        AnySplat
        & \xmark
        & \cmark
        & 23.39/0.80/0.14
        & 663K
        & 25.87/0.85/0.12
        & 18.85/0.45/0.36
        & 2850K
        & 19.18/0.51/0.32 \\

        DepthSplat-B
        & \xmark
        & \xmark
        & 25.83/0.90/0.11
        & 762K
        & 27.10/0.91/0.10
        & 16.44/0.40/0.50
        & 3175K
        & 19.16/0.52/0.38 \\

        \midrule

        MonoGS
        & \cmark
        & \cmark
        & 23.22/0.79/0.29
        & 6K
        & 22.71/0.79/0.27
        & 14.35/0.30/0.71
        & 10K
        & 14.21/0.29/0.68 \\

        \midrule

        CUT3R
        & \cmark
        & \cmark
        & 14.96/0.47/0.50
        & 762K
        & 14.62/0.41/0.57
        & 12.80/0.15/0.69
        & 3175K
        & 13.46/0.16/0.70 \\

        StreamSplat
        & \cmark
        & \cmark
        & --
        & 762K
        & 21.01/0.71/0.27
        & --
        & 3147K
        & 15.66/0.32/0.52 \\

        Spann3R
        & \cmark
        & \cmark
        & 15.06/0.44/0.49
        & 762K
        & 13.68/0.38/0.56
        & 11.98/0.10/0.69
        & 3175K
        & 12.15/0.11/0.70 \\

        Point3R
        & \cmark
        & \cmark
        & 15.11/0.45/0.44
        & 762K
        & 13.35/0.38/0.51
        & 11.38/0.10/0.69
        & 3175K
        & 11.14/0.10/0.71 \\

        StreamVGGT
        & \cmark
        & \cmark
        & 15.14/0.42/0.56
        & 762K
        & 14.65/0.34/0.64
        & 11.02/0.08/0.77
        & 3175K
        & 10.81/0.06/0.78 \\

        \textbf{S2Tok (Ours)}
        & \cmark
        & \cmark
        & \textbf{23.86/0.77/0.24}
        & \textbf{17K}
        & \textbf{25.03/0.79/0.22}
        & \textbf{18.00/0.40/0.58}
        & \textbf{82K}
        & \textbf{19.16/0.44/0.52} \\

        \bottomrule[1.3pt]
    \end{tabular}%
    }
\end{table*}

\paragraph{Mip-NeRF 360.} 
Tab.~\ref{tab:re10k_mipnerf360_comparison} and Tab.~\ref{tab:mipnerf360_comparison-full} evaluate transfer to scenes outside the training datasets. S2Tok achieves higher PSNR than the compared feed-forward streaming methods in the reported settings, using 43K and 82K Gaussians for the 12-view and 50-view evaluations, respectively. Its LPIPS results remain less competitive in some comparisons. Fig.~\ref{fig:mipnerf_causal} and Fig.~\ref{fig:mipnerf_final} provide qualitative examples of the decoded scenes.

\paragraph{ScanNet++.}
On ScanNet++ (Tab.~\ref{tab:scannetpp_comparison-full}), S2Tok has the best \textbf{Final} scores and the fewest Gaussians among the compared feed-forward streaming methods. Under \textbf{Intermediate} evaluation, StreamSplat leads on all three metrics at 50 views and on SSIM at 12 views. Fig.~\ref{fig:scannet_causal} (top row) illustrates challenging geometry predictions; improving these remains future work. More results are in Sec.~\ref{sec:moreresults}.

\begingroup
\emergencystretch=1em
\paragraph{Reconstruction across updates.} At 50 views, Final PSNR is lower than Intermediate PSNR on DL3DV (18.00 vs. 19.11), RealEstate10K (23.62 vs. 25.04), and ScanNet++ (20.54 vs. 22.63). Continued state updates therefore do not consistently preserve earlier rendering fidelity. Although the evolving scene state remains directly decodable, these aggregate results do not establish monotonic improvement with additional observations (Sec.~\ref{sec:limitation}).\par
\endgroup

\subsection{Ablation Study}
Table~\ref{tab:dl3dv-inference-ablation} reports ablations. Removing $\mathrm{RoPE_{3D}}$ or the scale token reduces rendering quality. Admitting all frame tokens increases the Gaussian count from 66K to 518K and Intermediate PSNR from 19.11 to 20.39, but reduces Final PSNR from 18.00 to 17.50. Removing SCST marginally improves rendering quality but nearly doubles the Gaussian count and associated storage. At a matched average Gaussian count of 66K, however, the full model achieves better results on all reported Final and Intermediate rendering metrics, supporting a modest quality benefit from SCST at the same Gaussian budget.



\input{figures/fig_qual_scannet_final.tex}
\input{figures/fig_qual_mipnerf_causal.tex}
\begin{table}[htbp]
    \centering
    \caption{Ablation experiments on the 50-view DL3DV evaluation set.}
    \label{tab:dl3dv-inference-ablation}

    \setlength{\tabcolsep}{3.2pt}
    \renewcommand{\arraystretch}{1.05}
    \scriptsize

    \resizebox{\linewidth}{!}{%
        \begin{tabular}{@{}lccc|ccc|r@{}}
            \toprule
            \multirow{2}{*}{\textbf{Variant}}
            & \multicolumn{3}{c|}{\textbf{Final}}
            & \multicolumn{3}{c|}{\textbf{Intermediate}}
            & \multirow{2}{*}{%
                \makecell{\textbf{Avg.}\\\textbf{\#GS}}
            }
            \\
            \cmidrule(lr){2-4}
            \cmidrule(lr){5-7}
            & \textbf{PSNR}$\uparrow$
            & \textbf{SSIM}$\uparrow$
            & \textbf{LPIPS}$\downarrow$
            & \textbf{PSNR}$\uparrow$
            & \textbf{SSIM}$\uparrow$
            & \textbf{LPIPS}$\downarrow$
            &
            \\
            \midrule

            w/o $\mathrm{RoPE}_{\mathrm{3D}}$
            & 17.74 & 0.4117 & 0.5166
            & 18.85 & 0.4452 & 0.4794
            & 65K
            \\

            w/o Admission
            & 17.50 & 0.4076 & 0.5476
            & 20.39 & 0.5087 & 0.4149
            & 518K
            \\

            w/o SCST
            & 18.13 & 0.4281 & 0.4997
            & 19.67 & 0.4757 & 0.4463
            & 120K
            \\

            w/o SCST(matched budget)
            & 17.90 & 0.4216 & 0.5112
            & 19.03 & 0.4541 & 0.4730
            & 66K
            \\

            w/o scale token $s$
            & 17.89 & 0.4214 & 0.5126
            & 19.01 & 0.4543 & 0.4765
            & 66K
            \\

            \textbf{Full Model}
            & \textbf{18.00}
            & \textbf{0.4236}
            & \textbf{0.5071}
            & \textbf{19.11}
            & \textbf{0.4571}
            & \textbf{0.4718}
            & \textbf{66K}
            \\

            \bottomrule
        \end{tabular}%
    }
\end{table}

\FloatBarrier
\section{Conclusion}
\label{sec:conclusion}
We propose \textbf{S2Tok}, a feed-forward framework that maintains a size-adaptive latent scene state for 3D Gaussian reconstruction from uncalibrated image streams. By separating observation integration from selective state expansion, S2Tok incorporates incoming frames into a shared token representation and jointly decodes the accumulated scene without retaining past images at inference time. Experiments across four benchmarks demonstrate competitive rendering performance with compact Gaussian outputs. These results support persistent spatial tokens as a practical representation for streaming scene reconstruction. Improving fidelity retention over continued updates and controlling state growth over longer streams remain important directions for future work.

\clearpage
\bibliographystyle{ACM-Reference-Format}
\bibliography{iclr2027_conference}

@article{zipsplat,
  title={ZipSplat: Fewer Gaussians, Better Splats},
  author={Veicht, Alexander and Hong, Sunghwan and Bar{\'a}th, D{\'a}niel and Pollefeys, Marc},
  journal={arXiv preprint arXiv:2606.05102},
  year={2026}
}

@article{tokengs,
  title={Tokengs: Decoupling 3d gaussian prediction from pixels with learnable tokens},
  author={Ren, Jiawei and Tyszkiewicz, Michal Jan and Huang, Jiahui and Gojcic, Zan},
  journal={arXiv preprint arXiv:2604.15239},
  year={2026}
}

@article{ros-cam,
  title={RGB-Only Supervised Camera Parameter Optimization in Dynamic Scenes},
  author={Li, Fang and Zhang, Hao and Ahuja, Narendra},
  journal={Advances in Neural Information Processing Systems},
  volume={38},
  pages={58160--58193},
  year={2026}
}

@inproceedings{streamgs,
  title={Streamgs: Online generalizable gaussian splatting reconstruction for unposed image streams},
  author={Li, Yang and Wang, Jinglu and Chu, Lei and Li, Xiao and Kao, Shiu-hong and Chen, Ying-Cong and Lu, Yan},
  booktitle={2025 IEEE/CVF International Conference on Computer Vision (ICCV)},
  pages={25841--25850},
  year={2025},
  organization={IEEE}
}

@inproceedings{streamvggt,
  title={Streaming visual geometry transformer},
  author={Zhuo, Dong and Zheng, Wenzhao and Guo, Jiahe and Wu, Yuqi and Zhou, Jie and Lu, Jiwen},
  booktitle={International Conference on Learning Representations},
  volume={2026},
  pages={88055--88072},
  year={2026}
}

@inproceedings{streamsplat,
  title={Streamsplat: Towards online dynamic 3d reconstruction from uncalibrated video streams},
  author={Wu, Zike and Yan, Qi and Yi, Xuanyu and Wang, Lele and Liao, Renjie},
  booktitle={International Conference on Learning Representations},
  volume={2026},
  pages={123114--123137},
  year={2026}
}

@article{sc4dgs,
  title={Self-calibrating 4d novel view synthesis from monocular videos using gaussian splatting},
  author={Li, Fang and Zhang, Hao and Ahuja, Narendra},
  journal={arXiv preprint arXiv:2406.01042},
  year={2024}
}

@article{globalsplat,
  title={GlobalSplat: Efficient Feed-Forward 3D Gaussian Splatting via Global Scene Tokens},
  author={Itkin, Roni and Issachar, Noam and Keypur, Yehonatan and Chen, Xingyu and Chen, Anpei and Benaim, Sagie},
  journal={arXiv preprint arXiv:2604.15284},
  year={2026}
}

@inproceedings{cut3r,
  title={Continuous 3d perception model with persistent state},
  author={Wang, Qianqian and Zhang, Yifei and Holynski, Aleksander and Efros, Alexei A and Kanazawa, Angjoo},
  booktitle={2025 IEEE/CVF Conference on Computer Vision and Pattern Recognition (CVPR)},
  pages={10510--10522},
  year={2025},
  organization={IEEE}
}

@article{point3r,
  title={Point3r: Streaming 3d reconstruction with explicit spatial pointer memory},
  author={Wu, Yuqi and Zheng, Wenzhao and Zhou, Jie and Lu, Jiwen},
  journal={Advances in Neural Information Processing Systems},
  volume={38},
  pages={69675--69699},
  year={2026}
}

@inproceedings{point4cast,
author = {Liu, Xinhang and Miraldo, Pedro and Lohit, Suhas and Jiang, Huaizu and Sawada, Naoko and Tai, Yu-Wing and Tang, Chi-Keung and Chatterjee, Moitreya},
title = {{Point4Cast: Streaming Dynamic Scene Reconstruction and Forecasting}},
booktitle = {IEEE Conference on Computer Vision and Pattern Recognition (CVPR)},
year = 2026,
pages = {14600--14611},
month = jun,
url = {https://www.merl.com/publications/TR2026-077}
}

@inproceedings{pixelsplat,
  title={pixelsplat: 3d gaussian splats from image pairs for scalable generalizable 3d reconstruction},
  author={Charatan, David and Li, Sizhe Lester and Tagliasacchi, Andrea and Sitzmann, Vincent},
  booktitle={2024 IEEE/CVF Conference on Computer Vision and Pattern Recognition (CVPR)},
  pages={19457--19467},
  year={2024},
  organization={IEEE}
}

@inproceedings{depthsplat,
  title={Depthsplat: Connecting gaussian splatting and depth},
  author={Xu, Haofei and Peng, Songyou and Wang, Fangjinhua and Blum, Hermann and Barath, Daniel and Geiger, Andreas and Pollefeys, Marc},
  booktitle={2025 IEEE/CVF Conference on Computer Vision and Pattern Recognition (CVPR)},
  pages={16453--16463},
  year={2025},
  organization={IEEE}
}

@inproceedings{dust3r,
  title={Dust3r: Geometric 3d vision made easy},
  author={Wang, Shuzhe and Leroy, Vincent and Cabon, Yohann and Chidlovskii, Boris and Revaud, Jerome},
  booktitle={2024 IEEE/CVF Conference on Computer Vision and Pattern Recognition (CVPR)},
  pages={20697--20709},
  year={2024},
  organization={IEEE}
}

@inproceedings{vggt,
  title={Vggt: Visual geometry grounded transformer},
  author={Wang, Jianyuan and Chen, Minghao and Karaev, Nikita and Vedaldi, Andrea and Rupprecht, Christian and Novotny, David},
  booktitle={2025 IEEE/CVF Conference on Computer Vision and Pattern Recognition (CVPR)},
  pages={5294--5306},
  year={2025},
  organization={IEEE}
}

@article{da3,
  title={Depth anything 3: Recovering the visual space from any views},
  author={Lin, Haotong and Chen, Sili and Liew, Junhao and Chen, Donny Y and Li, Zhenyu and Shi, Guang and Feng, Jiashi and Kang, Bingyi},
  journal={arXiv preprint arXiv:2511.10647},
  year={2025}
}

@inproceedings{limr,
  title={Learning implicit representation for reconstructing articulated objects},
  author={Zhang, Hao and Li, Fang and Rawlekar, Samyak and Ahuja, Narendra},
  booktitle={International Conference on Learning Representations},
  volume={2024},
  pages={51620--51641},
  year={2024}
}

@article{s3o,
  title={S3o: A dual-phase approach for reconstructing dynamic shape and skeleton of articulated objects from single monocular video},
  author={Zhang, Hao and Li, Fang and Rawlekar, Samyak and Ahuja, Narendra},
  journal={arXiv preprint arXiv:2405.12607},
  year={2024}
}

@article{helix4d,
  title={Helix4D: Complex 4D Mesh Generation},
  author={Yenphraphai, Jiraphon and Chen, Jianqi and Wang, Jian and Qian, Gordon and Tulyakov, Sergey and Abdal, Rameen and Yeh, Raymond A and Wonka, Peter and Wang, Chaoyang},
  journal={arXiv preprint arXiv:2605.26109},
  year={2026}
}

@article{vit,
  title={An image is worth 16x16 words: Transformers for image recognition at scale},
  author={Dosovitskiy, Alexey and Beyer, Lucas and Kolesnikov, Alexander and Weissenborn, Dirk and Zhai, Xiaohua and Unterthiner, Thomas and Dehghani, Mostafa and Minderer, Matthias and Heigold, Georg and Gelly, Sylvain and others},
  journal={arXiv preprint arXiv:2010.11929},
  year={2020}
}

@article{rope,
  title={Enhanced transformer with rotary position embedding., 2021},
  author={Su, J and Lu, Y and Pan, S and Murtadha, A and Wen, B and Roformer, Y Liu},
  journal={DOI: https://doi. org/10.1016/j. neucom},
  year={2023}
}

@inproceedings{mvsplat,
  title={Mvsplat: Efficient 3d gaussian splatting from sparse multi-view images},
  author={Chen, Yuedong and Xu, Haofei and Zheng, Chuanxia and Zhuang, Bohan and Pollefeys, Marc and Geiger, Andreas and Cham, Tat-Jen and Cai, Jianfei},
  booktitle={European conference on computer vision},
  pages={370--386},
  year={2024},
  organization={Springer}
}

@inproceedings{noposplat,
  title={No pose, no problem: Surprisingly simple 3d gaussian splats from sparse unposed images},
  author={Ye, Botao and Liu, Sifei and Xu, Haofei and Li, Xueting and Pollefeys, Marc and Yang, Ming-Hsuan and Peng, Songyou},
  booktitle={International Conference on Learning Representations},
  volume={2025},
  pages={54009--54033},
  year={2025}
}

@inproceedings{incvggt,
  title={IncVGGT: Incremental VGGT for memory-bounded long-range 3d reconstruction},
  author={Fang, Keyu and Zhou, Changchun and Fu, Yuzhe and Li, Hai and Chen, Yiran},
  booktitle={International Conference on Learning Representations},
  volume={2026},
  pages={5609--5623},
  year={2026}
}

@inproceedings{lrm,
  title={Lrm: Large reconstruction model for single image to 3d},
  author={Hong, Yicong and Zhang, Kai and Gu, Jiuxiang and Bi, Sai and Zhou, Yang and Liu, Difan and Liu, Feng and Sunkavalli, Kalyan and Bui, Trung and Tan, Hao},
  booktitle={International Conference on Learning Representations},
  volume={2024},
  pages={50678--50702},
  year={2024}
}

@article{rel10k,
  title={Stereo magnification: Learning view synthesis using multiplane images},
  author={Zhou, Tinghui and Tucker, Richard and Flynn, John and Fyffe, Graham and Snavely, Noah},
  journal={arXiv preprint arXiv:1805.09817},
  year={2018}
}

@inproceedings{dl3dv,
  title={Dl3dv-10k: A large-scale scene dataset for deep learning-based 3d vision},
  author={Ling, Lu and Sheng, Yichen and Tu, Zhi and Zhao, Wentian and Xin, Cheng and Wan, Kun and Yu, Lantao and Guo, Qianyu and Yu, Zixun and Lu, Yawen and others},
  booktitle={2024 IEEE/CVF Conference on Computer Vision and Pattern Recognition (CVPR)},
  pages={22160--22169},
  year={2024},
  organization={IEEE}
}

@article{arkitscenes,
  title={Arkitscenes: A diverse real-world dataset for 3d indoor scene understanding using mobile rgb-d data},
  author={Baruch, Gilad and Chen, Zhuoyuan and Dehghan, Afshin and Dimry, Tal and Feigin, Yuri and Fu, Peter and Gebauer, Thomas and Joffe, Brandon and Kurz, Daniel and Schwartz, Arik and others},
  journal={arXiv preprint arXiv:2111.08897},
  year={2021}
}

@inproceedings{hypersim,
    author    = {Mike Roberts AND Jason Ramapuram AND Anurag Ranjan AND Atulit Kumar AND
                 Miguel Angel Bautista AND Nathan Paczan AND Russ Webb AND Joshua M. Susskind},
    title     = {{Hypersim}: {A} Photorealistic Synthetic Dataset for Holistic Indoor Scene Understanding},
    booktitle = {International Conference on Computer Vision (ICCV) 2021},
    year      = {2021}
}

@inproceedings{scannet++,
  title={Scannet++: A high-fidelity dataset of 3d indoor scenes},
  author={Yeshwanth, Chandan and Liu, Yueh-Cheng and Nie{\ss}ner, Matthias and Dai, Angela},
  booktitle={2023 IEEE/CVF International Conference on Computer Vision (ICCV)},
  pages={12--22},
  year={2023},
  organization={IEEE}
}

@inproceedings{matsuki2024gaussian,
  title={Gaussian splatting slam},
  author={Matsuki, Hidenobu and Murai, Riku and Kelly, Paul HJ and Davison, Andrew J},
  booktitle={2024 IEEE/CVF Conference on Computer Vision and Pattern Recognition (CVPR)},
  pages={18039--18048},
  year={2024},
  organization={IEEE}
}

@inproceedings{flashmono,
 author = {Zhang, Zicheng and Wu, Ke and Meng, Xiangting and Liu, Keyu and Zhao, Jieru and Ding, Wenchao},
 booktitle = {International Conference on Learning Representations},
 editor = {C. Vondrick and B. Hariharan and C. Raffel and L. Pinto and D. Yang and A. Faust},
 pages = {58439--58458},
 title = {Flash-Mono: Feed-Forward Accelerated Gaussian Splatting Monocular SLAM},
 url = {https://proceedings.iclr.cc/paper_files/paper/2026/file/5fcacc84d87c97bee5baecb3fb8f6e88-Paper-Conference.pdf},
 volume = {2026},
 year = {2026}
}

@inproceedings{mipnerf360,
  title={Mip-nerf 360: Unbounded anti-aliased neural radiance fields},
  author={Barron, Jonathan T and Mildenhall, Ben and Verbin, Dor and Srinivasan, Pratul P and Hedman, Peter},
  booktitle={2022 IEEE/CVF Conference on Computer Vision and Pattern Recognition (CVPR)},
  pages={5460--5469},
  year={2022},
  organization={IEEE}
}

@inproceedings{scst,
  title={Self-critical sequence training for image captioning},
  author={Rennie, Steven J and Marcheret, Etienne and Mroueh, Youssef and Ross, Jerret and Goel, Vaibhava},
  booktitle={2017 IEEE conference on computer vision and pattern recognition (CVPR)},
  pages={1179--1195},
  year={2017},
  organization={IEEE}
}

@inproceedings{uniquer,
  title={UniQueR: Unified Query-based Feedforward 3D Reconstruction},
  author={Peng, Chensheng and Herau, Quentin and Yang, Jiezhi and Xie, Yichen and Hu, Yihan and Zheng, Wenzhao and Strong, Matthew and Tomizuka, Masayoshi and Zhan, Wei},
  booktitle = {European Conference on Computer Vision (ECCV)},
  year={2026}
}

@inproceedings{zwicker2001surface,
  author    = {Zwicker, Matthias and Pfister, Hanspeter and
               van Baar, Jeroen and Gross, Markus},
  title     = {Surface Splatting},
  booktitle = {Proceedings of the 28th Annual Conference on
               Computer Graphics and Interactive Techniques},
  pages     = {371--378},
  year      = {2001},
  doi       = {10.1145/383259.383300}
}

@inproceedings{pfister2000surfels,
  author    = {Pfister, Hanspeter and Zwicker, Matthias and
               van Baar, Jeroen and Gross, Markus},
  title     = {Surfels: Surface Elements as Rendering Primitives},
  booktitle = {Proceedings of the 27th Annual Conference on
               Computer Graphics and Interactive Techniques},
  pages     = {335--342},
  year      = {2000},
  doi       = {10.1145/344779.344936}
}
\clearpage
\appendix
\section*{Appendix}
\label{sec: appendix}
For ease of navigation, we organize the appendix as follows:

\begin{tabular}{@{}p{0.09\linewidth}p{0.87\linewidth}@{}}
\hyperref[sec:admitpseudogt]
{\textbf{\ref*{sec:admitpseudogt}}}
&
Admission Pseudo-GT
\\[2pt]
\hyperref[sec:scale_gt]
{\textbf{\ref*{sec:scale_gt}}}
&
Normalization Scale Pseudo-GT
\\[2pt]
\hyperref[sec:weightedaverage]
{\textbf{\ref*{sec:weightedaverage}}}
&
Opacity-Weighted Geometric Median of Gaussian Centers
\\[2pt]
\hyperref[sec:datasets]
{\textbf{\ref*{sec:datasets}}}
&
Datasets
\\[2pt]
\hyperref[sec:implementation]
{\textbf{\ref*{sec:implementation}}}
&
Implementation Details
\\[2pt]
\hyperref[sec:architecture-details]
{\hspace{1em}\ref*{sec:architecture-details}}
&
Architecture Details
\\[1pt]
\hyperref[sec:colorskipbranch]
{\hspace{1em}\ref*{sec:colorskipbranch}}
&
Color-Skip Branch
\\[1pt]
\hyperref[sec:training-details]
{\hspace{1em}\ref*{sec:training-details}}
&
Training Details
\\[1pt]
\hyperref[sec:hyperparameters]
{\hspace{1em}\ref*{sec:hyperparameters}}
&
Hyperparameters
\\[1pt]
\hyperref[sec:evaluation-protocol]
{\hspace{1em}\ref*{sec:evaluation-protocol}}
&
Evaluation Protocol
\\[2pt]
\hyperref[sec:detailcomparison]
{\textbf{\ref*{sec:detailcomparison}}}
&
Detailed Comparison of Related 3D Reconstruction Methods
\\[2pt]
\hyperref[sec:moreresults]
{\textbf{\ref*{sec:moreresults}}}
&
More Results
\\[2pt]
\hyperref[sec:limitation]
{\textbf{\ref*{sec:limitation}}}
&
Limitations
\end{tabular}

\section{Admission Pseudo-GT}
\label{sec:admitpseudogt}

We generate pseudo-GT admission labels from per-frame depth maps, camera
intrinsics, and camera poses that are available only during training; at test
time the admission module operates on images alone. The pseudo-GT criterion is a
depth-reprojection novelty test that marks the image patches observing
geometrically unseen regions of the scene.

\paragraph{Pseudo-GT depth and cameras.}
Camera intrinsics $K_i$ and camera-to-reference poses $T_i$ are taken directly
from the dataset annotations (COLMAP or capture metadata); all poses are
expressed relative to the first frame of the clip. Only the depth maps are
predicted. We run DA3~\citep{da3} (the nested Giant-Large checkpoint) offline
over every training clip, conditioning it on the images together with the
dataset intrinsics and world-to-camera extrinsics, and requesting depth aligned
to the scale of the input extrinsics. The resulting depth maps therefore live
in the same metric gauge as the pose translations, which makes reprojected
depths directly comparable across frames and keeps the scale-normalization
targets below consistent with the geometry supervision. Clips whose length
exceeds the DA3 context budget of $150$ views are processed in round-robin
interleaved chunks (frame $i$ is assigned to chunk $i \bmod n_c$), so each
forward pass spans the full trajectory and the extrinsic-scale alignment stays
globally consistent across chunks. Depths are stored with log-space 8-bit
quantization whose range is set by the $0.1$/$99.9$ depth percentiles
($<\!1\%$ relative error); out-of-range pixels are marked invalid.

\paragraph{Depth-reprojection novelty.}
Let $D_i(\mathbf{u})$ denote the depth of pixel $\mathbf{u}$ in frame $i$, and
let $K_i$ and $T_i$ denote its camera intrinsics and camera-to-reference
transformation, respectively. For every valid pixel $\mathbf{u}$ in frame $i$,
we first unproject it into 3D:
\begin{equation}
    \mathbf{x}_i(\mathbf{u})
    =
    T_i
    \left[
    D_i(\mathbf{u}) K_i^{-1}
    \widetilde{\mathbf{u}}
    \right],
\end{equation}
where $\widetilde{\mathbf{u}}$ is the homogeneous image coordinate. We then
project $\mathbf{x}_i(\mathbf{u})$ into every previously observed frame $k<i$.
A pixel is considered covered by frame $k$ if its projection is inside the
image, lies in front of the camera, has valid reference depth, and satisfies
\begin{equation}
    \left|
    z_{i\rightarrow k}(\mathbf{u})
    -
    D_k\!\left(\mathbf{u}_{i\rightarrow k}\right)
    \right|
    \leq
    \epsilon_{\mathrm{geo}}\,
    D_k\!\left(\mathbf{u}_{i\rightarrow k}\right),
\end{equation}
where $\mathbf{u}_{i\rightarrow k}$ and $z_{i\rightarrow k}$ are the projected
pixel coordinate and depth in frame $k$. In our experiments, we use
$\epsilon_{\mathrm{geo}}=0.08$. Because the test is relative to the reference
depth, it is invariant to the global scale of the scene.

A pixel is covered if the above condition is satisfied in at least one previous frame. Pixels with invalid depth in the incoming frame are conservatively treated as covered to avoid admissions based on unreliable geometry. Since admitted tokens are not removed, this rule limits spurious state expansion, but can also exclude genuinely new content.
The geometric novelty ratio of patch $\mathcal{P}_{ij}$ is
\begin{equation}
    r^{\mathrm{geo}}_{ij}
    =
    \frac{1}{|\mathcal{P}_{ij}|}
    \sum_{\mathbf{u}\in\mathcal{P}_{ij}}
    \mathbbm{1}\!\left[
        \mathbf{u}\ \text{is not covered}
    \right],
\end{equation}
computed over non-overlapping $14{\times}14$ patches aligned with the token
grid. The patch is labeled as admitted if
$r^{\mathrm{geo}}_{ij}>\tau_{\mathrm{geo}}$, where $\tau_{\mathrm{geo}}=0.3$.
All patches of the first frame are admitted unconditionally.

\paragraph{Causality of the labels.}
Although DA3~\cite{da3} processes each training clip as a whole when producing the
teacher depth maps, the coverage comparison is restricted to the observed prefix: the
label for frame $i$ compares frame $i$ only against the observed prefix
$\{0,\dots,i-1\}$, never against future frames. This matches the online
inference protocol, in which the admission module sees each incoming frame given
only the past. The full-clip DA3 pass is a teacher-side convenience that
improves cross-frame depth consistency. The teacher depths may depend on future frames and are used only for offline supervision; inference uses the current observation and the carried state.

\paragraph{Prefix scale for normalization.}
The same precomputation also yields a per-step scene scale used to normalize
pose-translation and depth supervision. After consuming frames $0,\dots,i$, we
collect the 3D points unprojected from the \emph{admitted} patches' valid
pixels of all frames so far, subsampled with a pixel stride of $2$ and capped
at $2{\times}10^{6}$ accumulated points by uniform random rejection. The
prefix scale $\ell_i$ is the median distance of these points to the origin of the
frame-$0$ coordinate system, clamped to $[10^{-2}, 10^{3}]$. Because only
admitted (i.e., deduplicated) content contributes, $\ell_i$ tracks the scale of
the accumulated scene memory rather than that of any single view, and it is
consistent with the DA3 depth gauge by construction.

\section{Normalization Scale Pseudo-GT}
\label{sec:scale_gt}
\paragraph{Normalization scale.}
For each step $i$, we collect 3D points unprojected from the valid pixels of
the geometrically novel patches admitted from frames $0,\ldots,i$.
Specifically, let $\mathcal{A}^{\mathrm{geo}}_k$ be the set of patches
selected by the depth-reprojection criterion in frame $k$. The accumulated
point set is
\begin{equation}
    \mathcal{Q}_i
    =
    \bigcup_{k=0}^{i}
    \left\{
        T_k
        \left[
        D_k(\mathbf{u})K_k^{-1}\widetilde{\mathbf{u}}
        \right]
        \,\middle|\,
        \mathbf{u}\in\mathcal{A}^{\mathrm{geo}}_k,\;
        D_k(\mathbf{u})>0
    \right\}.
\end{equation}
We use only the depth-reprojection admissions
(Sec.~\ref{sec:admitpseudogt}) when constructing $\mathcal{Q}_i$.
This pins the normalization gauge to scene geometry and prevents the scale
from changing with the admission policy.

The pseudo-GT normalization scale is the median distance from the accumulated
points to the reference origin:
\begin{equation}
    \ell_i
    =
    \operatorname*{median}_{\mathbf{x}\in\mathcal{Q}_i}
    \|\mathbf{x}\|_2.
\end{equation}
For numerical stability, we clamp the resulting scale to
$[10^{-2},10^{3}]$. In practice, pixels are spatially subsampled when
constructing the scale point set, and the total number of points is capped to
bound memory consumption.

\paragraph{Supervision-side normalization.}
At step $i$, the pseudo-GT scale $\ell_i$ defines the coordinate gauge for
all supervision performed at that step. For each selected context or target
view $j$, let
\begin{equation}
    T_j =
    \begin{bmatrix}
        R_j & \mathbf{t}_j \\
        \mathbf{0}^{\mathsf T} & 1
    \end{bmatrix}
\end{equation}
denote its camera-to-reference transformation. We normalize its translation
and depth using the current scale $\ell_i$:
\begin{equation}
    \overline{T}^{(i)}_j
    =
    \begin{bmatrix}
        R_j & \mathbf{t}_j/\ell_i \\
        \mathbf{0}^{\mathsf T} & 1
    \end{bmatrix},
    \qquad
    \overline{D}^{(i)}_j(\mathbf{u})
    =
    \frac{D_j(\mathbf{u})}{\ell_i}.
\end{equation}
The rotation $R_j$, camera intrinsics $K_j$, and RGB image $I_j$ remain
unchanged. Consequently,
\begin{equation}
    \overline{T}^{(i)}_j
    \left[
        \overline{D}^{(i)}_j(\mathbf{u})
        K_j^{-1}\widetilde{\mathbf{u}}
    \right]
    =
    \frac{1}{\ell_i}
    T_j
    \left[
        D_j(\mathbf{u})
        K_j^{-1}\widetilde{\mathbf{u}}
    \right],
\end{equation}
so the normalized camera translations, scene geometry, and depths remain
geometrically consistent.

The scene representation at step $i$ is rendered using
$\overline{T}^{(i)}_j$ and $K_j$. The rendered RGB is supervised by the
original image $I_j$, while the rendered depth is supervised by
$\overline{D}^{(i)}_j$. All views selected at step $i$, including previously
observed context and target views, are normalized using the current
prefix scale $\ell_i$ rather than the scale at which they first appeared.
The camera translation used for rendering is likewise $\mathbf{t}_i/\ell_i$, while its
rotation target remains $R_i$.

Consistent with the normalization-scale prediction described in the main
text, the MLP scale head takes the SS-Transformer output $s'_i$ and predicts
the log normalization scale $\hat{\ell}_i$. It is supervised by the same
pseudo-GT scale $\ell_i$ through
\begin{equation}
    \mathcal{L}_{\mathrm{scale}}^{(i)}
    =
    \lambda_{\mathrm{scale}}
    \left(
        \hat{\ell}_i-\log\ell_i
    \right)^2.
\end{equation}

\section{Opacity-Weighted Geometric Median of Gaussian Centers}
\label{sec:weightedaverage}

Each latent spatial token is decoded into $h$ Gaussian candidates. To assign a single robust 3D coordinate to a spatial token for the scene-local attention, we aggregate the Gaussian centers using an opacity-weighted geometric median.

Let $\mu_{mj}\in\mathbb{R}^{3}$ and $\alpha_{mj}\in(0,1)$ denote the center and activated opacity of the $j$-th Gaussian decoded from spatial token $m$, respectively. We first clamp the Gaussian centers to the valid spatial range:
\begin{equation}
    \overline{\mu}_{mj}
    =
    \operatorname{clamp}(\mu_{mj},-10,10).
\end{equation}
The coordinate assigned to token $m$ is the minimizer
\begin{equation}
    \mathbf{x}_m
    =
    \arg\min_{\mathbf{x}\in\mathbb{R}^{3}}
    \sum_{j=1}^{h}
    \alpha_{mj}
    \left\|
        \mathbf{x}-\overline{\mu}_{mj}
    \right\|_2.
\end{equation}
Unlike an arithmetic mean, this geometric median is robust to Gaussian candidates whose centers are spatial outliers.

We approximate the solution using the differentiable Weiszfeld iteration. It is initialized by the opacity-weighted arithmetic mean:
\begin{equation}
    \mathbf{x}^{(0)}_m
    =
    \frac{
        \sum_{j=1}^{h}\alpha_{mj}\overline{\mu}_{mj}
    }{
        \sum_{j=1}^{h}\alpha_{mj}+\varepsilon
    }.
\end{equation}
At iteration $t$, we compute
\begin{equation}
    d^{(t)}_{mj}
    =
    \max\left(
        \left\|
            \overline{\mu}_{mj}-\mathbf{x}^{(t)}_m
        \right\|_2,
        \varepsilon
    \right),
\end{equation}
followed by
\begin{equation}
    \mathbf{x}^{(t+1)}_m
    =
    \frac{
        \sum_{j=1}^{h}
        \frac{\alpha_{mj}}{d^{(t)}_{mj}}
        \overline{\mu}_{mj}
    }{
        \sum_{j=1}^{h}
        \frac{\alpha_{mj}}{d^{(t)}_{mj}}
        +\varepsilon
    }.
\end{equation}
We use eight iterations and $\varepsilon=10^{-6}$. If the total opacity of a token is smaller than $10^{-3}$, we use uniform weights to avoid an ill-defined coordinate.

Finally, we robustly estimate the center $\mathbf{c}$ of all token coordinates, using each token's total opacity
$m_m=\sum_j\alpha_{mj}$ as its weight, and use
\begin{equation}
    \widetilde{\mathbf{x}}_m=\mathbf{x}_m-\mathbf{c}
\end{equation}
as the centered 3D coordinate supplied to $\mathrm{RoPE_{3D}}$. This aggregation remains differentiable with respect to both Gaussian centers and opacities, allowing the subsequent reconstruction loss to supervise the spatial coordinates used by scene-local attention.

\section{Implementation Details}
\label{sec:implementation}

\subsection{Architecture Details}
\label{sec:architecture-details}

We use the weights of the pre-trained ZipSplat ViT-G/14 with a patch size of $w=14$ to initialize our model. Notably, ZipSplat is trained on the same datasets as S2Tok. We initialize the local scene attention with the same weights as the local frame attention. Our encoder contains 13 self-attention Transformer blocks to process only the incoming frame with a feature dimension of $D=1536$. Our SS-Transformer contains 27 blocks, each composed of a global attention layer, a local scene attention layer, and a local frame attention layer.

Features from the 7th, 17th, and 27th blocks in the SS-Transformer are passed to the hierarchical decoder. Each spatial token is decoded into $h=32$ Gaussians with degree-1 spherical harmonics.

The admission module first projects the frame and spatial tokens to 256
dimensions, followed by one cross-attention layer, one self-attention layer,
and an MLP classifier. At inference, a frame token is admitted when
$\sigma(\pi)>\tau$, where $\tau=0.5$.

\subsection{Color-Skip Branch}
\label{sec:colorskipbranch}
We additionally use a 128-dimensional color-skip branch adapted to our streaming setting from the color-skip branch in ZipSplat.

Following ZipSplat~\citep{zipsplat}, we attach a $128$-dimensional color-skip
branch that routes high-frequency appearance around the backbone directly to
the Gaussian head. In ZipSplat this branch embeds \emph{all} context views at
once and lets the retained tokens cross-attend to the color tokens of every
view; this is incompatible with streaming, where past frames are discarded. We
therefore restructure the branch around a per-token \emph{color bank} that is
maintained in lockstep with the scene memory. When frame $i$ arrives, a
lightweight patch embedding maps it to per-patch color tokens
$\mathbf{c}_i \in \mathbb{R}^{T\times 128}$, and only the newly admitted
patches query a single cross-attention block (operating entirely in the
$128$-dimensional color space) whose keys and values are the color tokens of
the \emph{current frame only}. The resulting color features are appended
to the color bank, row-aligned with the newly admitted scene-memory tokens.
The two memories are updated asymmetrically. Scene-memory tokens are
re-contextualized by the backbone at every step, so the geometric feature of
an old token continues to evolve as new observations arrive. Color-bank
entries, in contrast, are \emph{write-once}: the color feature of a token is
computed exactly once, from the frame that admitted it, and is retained frozen
thereafter---it is never recomputed, never passes through the backbone or
decoder, and receives no recurrent update. At every step, the Gaussian head
consumes the concatenation of each token's current decoded feature with its
frozen color feature, for old and newly admitted tokens alike. 
The branch computes new color features from the current frame and reuses previously stored color-bank entries without revisiting past images. It supplies admission-time appearance features to the Gaussian head alongside the recurrently updated scene features.


\subsection{Training Details}
\label{sec:training-details}
Training consists of three stages.
In the first stage, we train the model together with our proposed admission
module, but use the depth-reprojection novelty oracle
(Appendix~\ref{sec:admitpseudogt}) to select from new frame tokens. This
provides a stable admission policy for training the other modules, including
the Encoder, SS-Transformer, Hierarchical Decoder, and Gaussian Head. The
admission module is trained concurrently by distilling the oracle decisions,
but its outputs are not used for frame-token selection in this stage. We train for
100 epochs in the first stage on 128 NVIDIA A100 80GB GPUs, with the overall
training loss:
\begin{equation}
    \mathcal{L} = \mathcal{L}_{\mathrm{render}} + \mathcal{L}_{\mathrm{chamfer}} + \mathcal{L}_{\mathrm{scale}} + \mathcal{L}_{\mathrm{bce}}
\end{equation}

In the second stage, we enable SCST training and use the outputs of the
admission module itself to select frame tokens. This stage lets the admission
module learn directly from reconstruction performance, and lets the other
modules dynamically adapt to the newly learned policy. We train for 100 epochs in
the second stage on 128 NVIDIA A100 80GB GPUs, with the overall training loss:
\begin{equation}
    \mathcal{L} = \mathcal{L}_{\mathrm{render}} + \mathcal{L}_{\mathrm{chamfer}} + \mathcal{L}_{\mathrm{scale}} + \mathcal{L}_{\mathrm{scst}}
\end{equation}
In the first two stages, we train S2Tok using context sequences whose lengths
are uniformly sampled from $[12,16]$.

In the final stage, we freeze the \emph{parameters} of the admission module
and train all remaining components of S2Tok for 74 epochs on 64 NVIDIA B200
GPUs. 
This stage adapts the remaining trainable components to deterministic admission while holding the admission-module parameters fixed.
To be precise, only the admission-module parameters are frozen: the encoder
and SS-Transformer features supplied to the module continue to be trained, so
the admission probabilities are recomputed online at every step from the
current features rather than replayed from precomputed masks. In this stage,
admission is no longer Bernoulli-sampled; each patch is admitted
deterministically when its admission probability exceeds $0.5$, which is
exactly the decision rule used at inference. The module inputs are detached,
so no gradient flows from the admission decision into the backbone.
Admission decisions can still change as the backbone features evolve. This stage therefore trains the remaining components under the deterministic admission rule used at inference, rather than the Bernoulli-sampled rule used in stage two.
We train with context sequences whose
lengths are uniformly sampled from $[12,24]$, with the overall training loss:
\begin{equation}
    \mathcal{L} = \mathcal{L}_{\mathrm{render}} + \mathcal{L}_{\mathrm{chamfer}} + \mathcal{L}_{\mathrm{scale}}
\end{equation}

\paragraph{Timestep-indexed SCST objective.}
We now give the exact per-timestep form of the SCST objective in
Sec.~\ref{sec:loss}. For each sequence, we unroll the deterministic
(baseline) rollout and the sampled rollout of Sec.~\ref{sec:loss} over the
full context sequence of length $V$ with shared weights. At step $i$, the
reconstruction error of a rollout is evaluated without gradient by rendering
\emph{all} views observed so far (context views $0{:}i$ and target views
$0{:}i{-}1$) from the accumulated scene:
\begin{equation}
e_i = \mathcal{L}_{\mathrm{rgb},i} + 0.05\,\mathcal{L}_{\mathrm{lpips},i}
      + 0.01\,\mathcal{L}_{\mathrm{depth},i},
\end{equation}
yielding $e^{b}_i$ for the sampled rollout and $e^{d}_i$ for the
deterministic baseline. The self-critical advantage $A$ of
Sec.~\ref{sec:loss} is instantiated at each step as a reward-to-go difference:
\begin{equation}
A_i = \sum_{j \ge i} e^{b}_j - \sum_{j \ge i} e^{d}_j,
\qquad i = 1,\dots,V{-}1,
\end{equation}
where step $0$ admits all tokens unconditionally and carries no decision.
The running scale $\hat\sigma_A$ is an exponential moving average of the
batch standard deviation of the advantages,
$\hat\sigma_A \leftarrow 0.99\,\hat\sigma_A +
0.01\operatorname{std}(\{A_i\})$, clamped to $\ge 10^{-6}$.

Let $\mathbf{a}^{b}_i$ denote the sampled admission mask over the $N_p$
candidate tokens of frame $i$, and $\mathbf{p}_i = \sigma(\symbfit{\pi}_i)$
the corresponding admission probabilities. The normalized log-probability
$\Theta$ decomposes per step as
\begin{equation}
\Theta_i(\mathbf{a}^{b}_i)
= \frac{1}{N_p}\log P_{\theta}\!\left(\mathbf{a}^{b}_i\right)
= \frac{1}{N_p} \sum_{p=1}^{N_p}
\Big[ a^{b}_{i,p} \log p_{i,p}
      + \big(1 - a^{b}_{i,p}\big) \log \big(1 - p_{i,p}\big) \Big],
\end{equation}
so each cumulative advantage $A_i$ weights only the log-probabilities of the
Bernoulli decisions taken at step $i$. The full SCST loss averages over the
$V{-}1$ decision steps:
\begin{equation}
\mathcal{L}_{scst} = \frac{1}{V-1} \sum_{i=1}^{V-1}
\left[
\frac{\operatorname{sg}(A_i)}{\hat\sigma_A}\,\Theta_i(\mathbf{a}^{b}_i)
+ \beta_{\mathrm{KL}}\,
  D_{\mathrm{KL}}\!\left(Q_i \,\Vert\, P_{\theta,i}\right)
\right],
\end{equation}
which reduces to Eq.~\ref{eq:scst} for a single decision step. Here
$Q_i$ is the label-smoothed oracle admission distribution: token $p$
receives target probability $0.95$ if the depth-reprojection oracle
(Appendix~\ref{sec:admitpseudogt}) admits it and $0.05$ otherwise, and
$D_{\mathrm{KL}}(Q_i \Vert P_{\theta,i})$ is the mean Bernoulli KL divergence
over the $N_p$ tokens. The weight $\beta_{\mathrm{KL}}$ is annealed linearly
from $1.0$ to a floor of $0.1$ over the first 500 iterations of the second
stage and held at $0.1$ thereafter. The token-mean normalization in
$\Theta_i$ and the average over decision steps rescale the estimator by
positive constants, leaving the policy gradient unbiased. Gradients from
$\mathcal{L}_{scst}$ reach only the admission module; its inputs are detached
from the backbone.

In each stage, we use AdamW with a base learning rate of $10^{-4}$, weight
decay $0.05$, and $(\beta_1,\beta_2)=(0.9,0.95)$. The encoder and aggregator
use a $0.1\times$ learning-rate multiplier. Warm-up covers the first 5\% of
each stage's optimizer steps (4{,}780 steps in the final stage), followed by
cosine decay; we apply gradient clipping at $1.0$ and gradient accumulation
over four micro-batches. Finally, we use full backpropagation through the
recurrent state ($\texttt{tbptt\_k}=0$), so gradients are not truncated within
a training sequence.

\subsection{Hyperparameters}
\label{sec:hyperparameters}
We set
$\lambda_{\mathrm{lpips}}=0.05$,
$\lambda_{\mathrm{depth}}=0.01$,
$\lambda^{\mathrm{solo}}_{\mathrm{render}}=0.5$,
$\lambda_{\mathrm{chamfer}}=0.05$,
$\lambda^{\mathrm{solo}}_{\mathrm{chamfer}}=0.05$, and
$\lambda_{\mathrm{scale}}=1$.
For the depth-reprojection novelty teacher, we use a relative depth tolerance
of $\epsilon_{\mathrm{geo}}=0.08$ and admit a patch when its novel-pixel
fraction exceeds $0.3$.
Because the scene subsets differ between the 12-view and 50-view evaluations for some datasets, comparisons across these settings should not be interpreted as a controlled sequence-length ablation.

\subsection{Evaluation Protocol}
\label{sec:evaluation-protocol}

\begingroup
\emergencystretch=2em
For the 12-view setting, we evaluate on 8 DL3DV, 8 RealEstate10K, 9
Mip-NeRF~360, and 8 ScanNet++ scenes. For the 50-view setting, we evaluate on
all held-out evaluation sets of DL3DV, RealEstate10K, and ScanNet++, and all scenes of Mip-NeRF~360. Context
views are sampled in temporal order, with one target view between
each pair of consecutive context views. In \textbf{Intermediate} evaluation, target
$t_{k-1}$ is rendered after observing context views $c_0,\ldots,c_k$;
in \textbf{Final} evaluation, all targets are rendered after observing the complete context stream.\par
\endgroup

\section{Datasets}
\label{sec:datasets}
\paragraph{Training datasets.}
We train our model on only two datasets: RealEstate10K~\citep{rel10k} and DL3DV~\citep{dl3dv}. Specifically, we use their training sets and keep the held-out evaluation sets for testing.



\paragraph{Testing datasets.}
We test our model on both the in-distribution RealEstate10K~\citep{rel10k} and DL3DV~\citep{dl3dv} evaluation sets and the out-of-distribution datasets Mip-NeRF 360~\citep{mipnerf360} and ScanNet++~\citep{scannet++}. In addition, we evaluate S2Tok on some in-the-wild videos and online videos.

\paragraph{Preprocessing and streaming sampling.}
All images are resized with the shorter side set to $252$ pixels, and their spatial dimensions are made divisible by the ViT patch size. Camera intrinsics are transformed consistently with image resizing and cropping. Invalid depth values and depths larger than $500$ are excluded from all geometry-dependent supervision.

During training, we sample an ordered stream containing between $12$ and $24$ context frames. The first sampled frame defines the reference coordinate system, and the remaining frames preserve their temporal order.

\section{Detailed Comparison of Related 3D Reconstruction Methods}
\label{sec:detailcomparison}
A more detailed method comparison is shown in Tab.~\ref{tab:method-comparison-detail}. Unlike the most closely related baselines---CUT3R, Point3R, and Point4Cast---which are trained on many datasets, S2Tok is trained on only 2 public datasets and demonstrates good zero-shot ability as shown in Sec.~\ref{sec:experiments}.

\begin{table*}[t]
\caption{Detailed Comparison of Related Feed-forward 3D Reconstruction Methods. Each method is shown separately. Training datasets refer to the datasets used for training or fine-tuning the reported model.}
\label{tab:method-comparison-detail}

\centering
\scriptsize
\setlength{\tabcolsep}{2pt}
\renewcommand{\arraystretch}{1.1}

\renewcommand{\tabularxcolumn}[1]{m{#1}}

\begin{tabularx}{\textwidth}{
    @{}
    c
    c
    c
    c
    c
    c
    c
    >{\centering\arraybackslash}X
    @{}
}
\toprule[1.3pt]

& \multicolumn{3}{c}{\textbf{Streaming Setting}}
& & & & \\
\cmidrule(lr){2-4}

\textbf{Method}
& \textbf{Streaming}
& \makecell{
    \textbf{Latent Spatial}\\
    \textbf{Token/Memory}
}
& \makecell{
    \textbf{Cache/Window-}\\
    \underline{\texttt{\textbf{free}}}
}
& \makecell{
    \textbf{Pixel-}\underline{\texttt{\textbf{free}}}\\
    \textbf{Prediction}
}
& \makecell{
    \textbf{Reconstruction}\\
    \textbf{Kernel}
}
& \makecell{
    \textbf{Pose-}\\
    \textbf{free}
}
& \makecell{
    \textbf{Training}\\
    \textbf{Dataset(s)}
}
\\
\midrule

AnySplat
& \xmark
& \xmark
& \makecell{non-\\streaming}
& \xmark
& GS
& \cmark
& Hypersim, ARKitScenes, BlendedMVS, ScanNet++,
  CO3D-v2, Objaverse, Unreal4K, WildRGB-D, and DL3DV
\\
\midrule

Spann3R
& \cmark
& \cmark
& \xmark
& \xmark
& Point
& \cmark
& Habitat, ScanNet, ScanNet++, ARKitScenes,
  BlendedMVS, and CO3D-v2\textsuperscript{\dag}
\\
\midrule

StreamVGGT
& \cmark
& \cmark
& \xmark
& \xmark
& Point
& \cmark
& CO3D-v2, BlendedMVS, ARKitScenes, MegaDepth,
  WildRGB-D, ScanNet, Hypersim, OmniObject3D,
  MVS-Synth, PointOdyssey, Virtual KITTI 2,
  Spring, and Waymo
\\
\midrule

StreamGS
& \makecell{pair-\\wise}
& \xmark
& \xmark
& \xmark
& GS
& \cmark
& RealEstate10K
\\
\midrule

StreamSplat
& \makecell{pair-\\wise}
& \xmark
& \xmark
& \xmark
& GS
& \cmark
& CO3D-v2, RealEstate10K, DAVIS, and YouTube-VOS
\\
\midrule

ZipSplat
& \xmark
& \cmark
& \makecell{non-\\streaming}
& \cmark
& GS
& \cmark
& RealEstate10K and DL3DV
\\
\midrule

TokenGS
& \xmark
& \cmark
& \makecell{non-\\streaming}
& \cmark
& GS
& \xmark
& RealEstate10K and DL3DV
\\
\midrule

GlobalSplat
& \xmark
& \cmark
& \makecell{non-\\streaming}
& \cmark
& GS
& \xmark
& RealEstate10K
\\
\midrule

CUT3R
& \cmark
& \cmark
& \cmark
& \xmark
& Point
& \cmark
& ARKitScenes, BlendedMVS, CO3D-v2, MegaDepth,
  ScanNet++, ScanNet, Waymo, WildRGB-D, MapFree,
  TartanAir, UnrealStereo4K, Virtual KITTI 2,
  3D Ken Burns, BEDLAM, COP3D, DL3DV,
  Dynamic Replica, EDEN, Hypersim, IRS,
  Matterport3D, MVImgNet, MVS-Synth,
  OmniObject3D, PointOdyssey, RealEstate10K,
  SmartPortraits, Spring, Synscapes, UASOL,
  UrbanSyn, and HOI4D
\\
\midrule

Point3R
& \cmark
& \cmark
& \cmark
& \xmark
& Point
& \cmark
& ARKitScenes, ScanNet, ScanNet++, CO3D-v2,
  WildRGB-D, OmniObject3D, Hypersim,
  BlendedMVS, MegaDepth, Waymo,
  Virtual KITTI 2, PointOdyssey, Spring,
  and MVS-Synth
\\
\midrule

Point4Cast
& \cmark
& \cmark
& \cmark
& \xmark
& Point
& \cmark
& Kubric, PointOdyssey, Stereo4D,
  Mixamo and BlenderKit assets
  (Blender-rendered synthetic dataset)
\\
\midrule

\textbf{S2Tok (Ours)}
& \cmark
& \cmark
& \cmark
& \cmark
& GS
& \cmark
& DL3DV, RealEstate10K
\\

\bottomrule[1.3pt]
\end{tabularx}

\vspace{3pt}

\begin{minipage}{\textwidth}
\centering
\scriptsize
\textsuperscript{\dag}The Spann3R paper lists these six datasets
as its training-data pool, but its limitations section states that
the reported model was trained on only four datasets without
identifying the exact subset.
\end{minipage}

\end{table*}

\section{More Results}
\label{sec:moreresults}
We put more results in this section to more concretely demonstrate the effectiveness of S2Tok.

Tab.~\ref{tab:re10k_comparison-full}, Tab.~\ref{tab:mipnerf360_comparison-full}, and Tab.~\ref{tab:scannetpp_comparison-full} provide detailed results on RealEstate10K, Mip-NeRF 360, and ScanNet++. S2Tok combines compact Gaussian outputs with strong PSNR results among the compared feed-forward streaming methods, but does not lead every metric or setting. In particular, StreamSplat achieves better results in the 50-view Intermediate evaluation on ScanNet++, and several methods obtain lower LPIPS in other comparisons. MonoGS obtains higher 50-view Final PSNR on RealEstate10K and ScanNet++, while using per-scene optimization and requiring substantially more reconstruction time in the timing experiment of Tab.~\ref{tab:dl3dv_comparison}.


Fig.~\ref{fig:mipnerf_final}, Fig.~\ref{fig:dl3dv_causal}, Fig.~\ref{fig:scannet_causal}, and Fig.~\ref{fig:scannet_final} illustrate qualitative differences between the evaluated reconstructions. In selected examples, S2Tok produces fewer isolated artifacts than several compared renderings, although blur and geometry errors remain. Architecturally, S2Tok predicts the current Gaussian scene jointly from its updated token representation in a shared reference frame, rather than treating previously decoded Gaussians as immutable outputs. This enables scene-level readout from the carried state, but does not by itself guarantee geometric accuracy or fidelity retention across updates. The comparisons also involve differences in supervision and rendering representations, so they do not isolate a single cause of the observed performance gaps.

\begin{figure}[htbp]
    \centering
    \includegraphics[width=\linewidth]{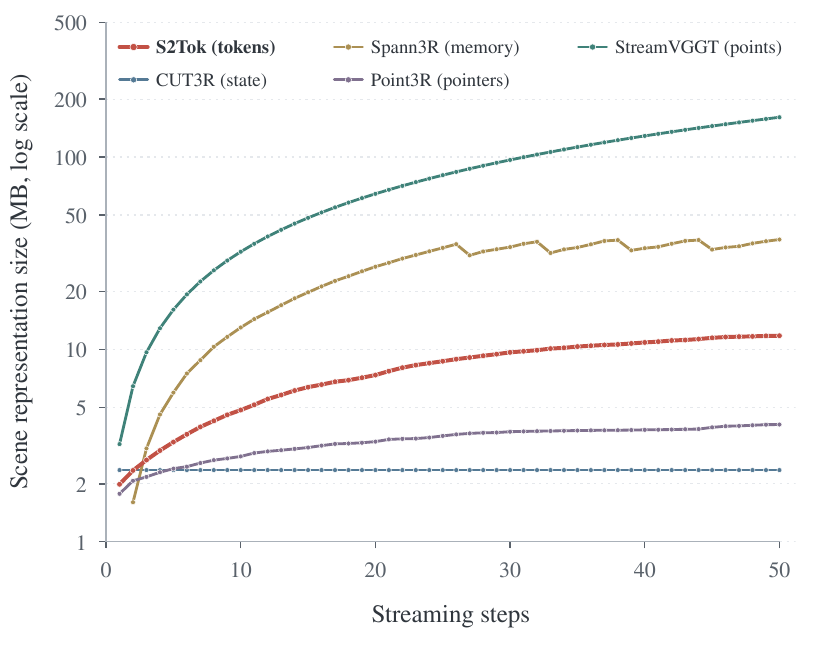}
    \caption{\textbf{Scene Representation Size Comparisons at Each Step.} We report the storage of the representation components identified in the legend, averaged over the evaluated sequences and shown on a logarithmic scale. The curves characterize storage growth over this finite evaluation window, rather than end-to-end GPU memory consumption.}
    \label{fig:scenesize}
\end{figure}

Fig.~\ref{fig:scenesize} reports the storage of the method-specific representation components identified in the legend, averaged over the evaluated 50-view sequences. The growth of S2Tok token storage becomes slower within this evaluation window, but the state continues to expand. This finite-horizon observation does not establish a bound on memory usage for longer streams. Moreover, these component-level storage measurements are distinct from end-to-end GPU memory consumption, for which Tab.~\ref{tab:dl3dv_comparison} reports peak usage.


\section{Limitations}
\label{sec:limitation}
S2Tok maintains a persistent scene state, but continued updates can reduce rendering fidelity on previously evaluated views, as reflected by the Final–Intermediate gaps in several settings. The model is trained on sequences of 12–24 context frames and evaluated quantitatively at 12 and 50 views; robust retention over longer streams remains unestablished. Admission controls which new tokens are added, but the current model does not remove or compress admitted tokens. State size and the cost of recurrent updating and scene decoding can therefore increase as observations accumulate. The method does not provide a fixed memory budget. The current framework is designed for static scenes. Extending the state representation and update mechanism to dynamic content remains future work.


\begin{table*}[htbp]
    \centering
    \caption{Quantitative Comparison on RealEstate10K. Each metric cell reports PSNR$\uparrow$ / SSIM$\uparrow$ / LPIPS$\downarrow$. \textbf{Final} evaluates all target views after observing the complete context stream, whereas \textbf{Intermediate} evaluates each target view when it first becomes renderable using only previously observed frames. $r{=}1$ denotes the default compression ratio of ZipSplat. $^{\dagger}$ adjusts the compression ratio to match our token budget. Notably, MonoGS is not feed-forward.}
    \label{tab:re10k_comparison-full}

    \renewcommand{\arraystretch}{1.05}
    \resizebox{\textwidth}{!}{%
    \begin{tabular}{
        @{}l
        @{\hspace{4pt}}c
        @{\hspace{4pt}}c
        @{\hspace{8pt}}ccc
        @{\hspace{8pt}}cc@{}
    }
        \toprule[1.3pt]

        \multirow{3}{*}{\textbf{Method}}
        & \multirow{3}{*}{\shortstack{\textbf{Stream-}\\\textbf{ing}}}
        & \multirow{3}{*}{\shortstack{\textbf{Cam-}\\\textbf{free}}}
        & \multicolumn{3}{c}{\textbf{Final}}
        & \multicolumn{2}{c}{\textbf{Intermediate}} \\
        \cmidrule(lr){4-6}
        \cmidrule(lr){7-8}

        & & &
        \multirow{2}{*}{\textbf{12 views}}
        & \multirow{2}{*}{\textbf{50 views}}
        & \textbf{Avg. \#GS/PT}
        & \multirow{2}{*}{\textbf{12 views}}
        & \multirow{2}{*}{\textbf{50 views}} \\

        & & & & & \textbf{12 / 50 views} & & \\

        \midrule

        ZipSplat$^{r{=}1}$
        & \xmark & \cmark
        & 27.65/0.87/0.13
        & 29.13/0.91/0.10
        & 124K/518K
        & 27.95/0.88/0.12
        & 29.29/0.91/0.10 \\

        ZipSplat$^{\dagger}$
        & \xmark & \cmark
        & 25.47/0.82/0.19
        & 26.49/0.86/0.17
        & 17K/21K
        & 26.26/0.84/0.17
        & 26.89/0.86/0.17 \\

        TokenGS
        & \xmark & \xmark
        & 32.00/0.94/0.15
        & 31.96/0.94/0.14
        & 262K/262K
        & 30.79/0.90/0.17
        & 32.47/0.94/0.14 \\

        GlobalSplat
        & \xmark & \xmark
        & 29.85/0.91/0.13
        & 29.04/0.89/0.16
        & 32K/32K
        & 29.23/0.89/0.14
        & 29.37/0.90/0.15 \\

        AnySplat
        & \xmark & \cmark
        & 23.39/0.80/0.14
        & 23.92/0.82/0.13
        & 663K/2305K
        & 25.87/0.85/0.12
        & 26.63/0.88/0.11 \\

        DepthSplat-B
        & \xmark & \xmark
        & 25.83/0.90/0.11
        & 25.44/0.89/0.12
        & 762K/3175K
        & 27.10/0.91/0.10
        & 26.74/0.93/0.09 \\

        \midrule

        MonoGS
        & \cmark & \cmark
        & 23.22/0.79/0.29
        & 23.73/0.83/0.24
        & 6K/8K
        & 22.71/0.79/0.27
        & 23.91/0.83/0.26 \\

        \midrule

        CUT3R
        & \cmark & \cmark
        & 14.96/0.47/0.50
        & 14.85/0.51/0.52
        & 762K/3175K
        & 14.62/0.41/0.57
        & 14.44/0.47/0.56 \\

        StreamSplat
        & \cmark & \cmark
        & --
        & --
        & 762K/3175K
        & 21.01/0.71/0.27
        & 22.26/0.74/0.27 \\

        Spann3R
        & \cmark & \cmark
        & 15.06/0.44/0.49
        & 15.57/0.43/0.53
        & 762K/3175K
        & 13.68/0.38/0.56
        & 13.14/0.34/0.62 \\

        Point3R
        & \cmark & \cmark
        & 15.11/0.45/0.44
        & 14.25/0.41/0.53
        & 762K/3175K
        & 13.35/0.38/0.51
        & 12.46/0.32/0.60 \\

        StreamVGGT
        & \cmark & \cmark
        & 15.14/0.42/0.56
        & 14.09/0.31/0.69
        & 762K/3175K
        & 14.65/0.34/0.64
        & 12.34/0.20/0.76 \\

        \textbf{S2Tok (Ours)}
        & \cmark & \cmark
        & \textbf{23.86/0.77/0.24}
        & \textbf{23.62/0.76/0.26}
        & \textbf{17K/21K}
        & \textbf{25.03/0.79/0.22}
        & \textbf{25.04/0.79/0.24} \\

        \bottomrule[1.3pt]
    \end{tabular}%
    }
\end{table*}

\begin{table*}[htbp]
    \centering
    \caption{Quantitative Comparison on Mip-NeRF 360. Each metric cell reports PSNR$\uparrow$ / SSIM$\uparrow$ / LPIPS$\downarrow$. \textbf{Final} evaluates all target views after observing the complete context stream, whereas \textbf{Intermediate} evaluates each target view when it first becomes renderable using only previously observed frames. $r{=}1$ denotes the default compression ratio of ZipSplat. ZipSplat$^{\dagger}$ adjusts its compression ratio to match the primitive budget of our method. Notably, MonoGS is not feed-forward.}
    \label{tab:mipnerf360_comparison-full}

    \renewcommand{\arraystretch}{1.05}
    \resizebox{\textwidth}{!}{%
    \begin{tabular}{
        @{}l
        @{\hspace{4pt}}c
        @{\hspace{4pt}}c
        @{\hspace{8pt}}ccc
        @{\hspace{8pt}}cc@{}
    }
        \toprule[1.3pt]

        \multirow{3}{*}{\textbf{Method}}
        & \multirow{3}{*}{\shortstack[c]{\textbf{Stream-}\\\textbf{ing}}}
        & \multirow{3}{*}{\shortstack[c]{\textbf{Cam-}\\\textbf{free}}}
        & \multicolumn{3}{c}{\textbf{Final}}
        & \multicolumn{2}{c}{\textbf{Intermediate}} \\
        \cmidrule(lr){4-6}
        \cmidrule(lr){7-8}

        & & &
        \multirow{2}{*}{\textbf{12 views}}
        & \multirow{2}{*}{\textbf{50 views}}
        & \textbf{Avg. \#GS/PT}
        & \multirow{2}{*}{\textbf{12 views}}
        & \multirow{2}{*}{\textbf{50 views}} \\

        & & & & & \textbf{12 / 50 views} & & \\

        \midrule

        ZipSplat$^{r{=}1}$
        & \xmark & \cmark
        & 20.71/0.50/0.40
        & 21.46/0.54/0.38
        & 124K/518K
        & 20.83/0.50/0.40
        & 21.39/0.53/0.38 \\

        ZipSplat$^{\dagger}$
        & \xmark & \cmark
        & 20.40/0.48/0.43
        & 20.66/0.50/0.43
        & 43K/82K
        & 20.48/0.49/0.43
        & 20.68/0.49/0.43 \\

        TokenGS
        & \xmark & \xmark
        & 22.55/0.67/0.42
        & 22.29/0.64/0.48
        & 262K/262K
        & 21.68/0.66/0.41
        & 22.88/0.67/0.43 \\

        GlobalSplat
        & \xmark & \xmark
        & 18.12/0.41/0.58
        & 17.25/0.39/0.65
        & 32K/32K
        & 17.50/0.40/0.57
        & 17.85/0.42/0.59 \\

        AnySplat
        & \xmark & \cmark
        & 17.15/0.43/0.36
        & 18.85/0.45/0.36
        & 719K/2850K
        & 16.52/0.45/0.34
        & 19.18/0.51/0.32 \\

        DepthSplat-B
        & \xmark & \xmark
        & 18.02/0.46/0.42
        & 16.44/0.40/0.50
        & 762K/3175K
        & 19.89/0.54/0.35
        & 19.16/0.52/0.38 \\

        \midrule

        MonoGS
        & \cmark & \cmark
        & 12.06/0.18/0.71
        & 14.35/0.30/0.71
        & 12K/10K
        & 13.86/0.26/0.66
        & 14.21/0.29/0.68 \\

        \midrule

        CUT3R
        & \cmark & \cmark
        & 12.95/0.20/0.65
        & 12.80/0.15/0.69
        & 762K/3175K
        & 12.19/0.17/0.67
        & 13.46/0.16/0.70 \\

        StreamSplat
        & \cmark & \cmark
        & --
        & --
        & 759K/3147K
        & 16.23/0.39/0.47
        & 15.66/0.32/0.52 \\

        Spann3R
        & \cmark & \cmark
        & 12.66/0.13/0.65
        & 11.98/0.10/0.69
        & 762K/3175K
        & 11.89/0.11/0.66
        & 12.15/0.11/0.70 \\

        Point3R
        & \cmark & \cmark
        & 12.44/0.15/0.63
        & 11.38/0.10/0.69
        & 762K/3175K
        & 11.90/0.14/0.64
        & 11.14/0.10/0.71 \\

        StreamVGGT
        & \cmark & \cmark
        & 11.86/0.14/0.72
        & 11.02/0.08/0.77
        & 762K/3175K
        & 10.70/0.09/0.75
        & 10.81/0.06/0.78 \\

        \textbf{S2Tok (Ours)}
        & \cmark & \cmark
        & \textbf{18.85/0.42/0.52}
        & \textbf{18.00/0.40/0.58}
        & \textbf{43K/82K}
        & \textbf{19.48/0.44/0.49}
        & \textbf{19.16/0.44/0.52} \\

        \bottomrule[1.3pt]
    \end{tabular}%
    }
\end{table*}

\begin{table*}[htbp]
    \centering
    \caption{Quantitative Comparison on ScanNet++. Each metric cell reports PSNR$\uparrow$ / SSIM$\uparrow$ / LPIPS$\downarrow$. \textbf{Final} evaluates target views after observing the complete context stream, while \textbf{Intermediate} evaluates each target view using only the context frames available at that step. ZipSplat$^{\dagger}$ uses the S2Tok-derived reduced-budget setting. Notably, MonoGS is not feed-forward.}
    \label{tab:scannetpp_comparison-full}

    \renewcommand{\arraystretch}{1.08}
    \resizebox{\textwidth}{!}{%
    \begin{tabular}{
        @{}l
        @{\hspace{4pt}}c
        @{\hspace{4pt}}c
        @{\hspace{8pt}}c
        @{\hspace{8pt}}c
        @{\hspace{8pt}}c
        @{\hspace{8pt}}c
        @{\hspace{8pt}}c@{}
    }
        \toprule[1.3pt]

        \multirow{3}{*}{\textbf{Method}}
        & \multirow{3}{*}{\shortstack[c]{\textbf{Stream-}\\\textbf{ing}}}
        & \multirow{3}{*}{\shortstack[c]{\textbf{Cam-}\\\textbf{free}}}
        & \multicolumn{3}{c}{\textbf{Final}}
        & \multicolumn{2}{c}{\textbf{Intermediate}} \\

        \cmidrule(lr){4-6}
        \cmidrule(lr){7-8}

        & & &
        \multirow{2}{*}{\textbf{12 views}}
        & \multirow{2}{*}{\textbf{50 views}}
        & \textbf{Avg. \#GS/PT}
        & \multirow{2}{*}{\textbf{12 views}}
        & \multirow{2}{*}{\textbf{50 views}} \\

        & & & & & \textbf{12 / 50 views} & & \\

        \midrule

        ZipSplat$^{r{=}1}$
        & \xmark & \cmark
        & 22.25/0.74/0.30
        & 21.46/0.73/0.35
        & 124K/518K
        & 23.21/0.76/0.27
        & 22.84/0.76/0.31 \\

        ZipSplat$^{\dagger}$
        & \xmark & \cmark
        & 21.99/0.74/0.31
        & 21.23/0.73/0.35
        & 41K/71K
        & 22.92/0.76/0.29
        & 22.55/0.76/0.32 \\

        TokenGS
        & \xmark & \xmark
        & 28.71/0.91/0.22
        & 28.99/0.91/0.25
        & 262K/262K
        & 29.26/0.92/0.20
        & 30.15/0.91/0.23 \\

        GlobalSplat
        & \xmark & \xmark
        & 22.63/0.79/0.34
        & 19.33/0.73/0.39
        & 32K/32K
        & 23.39/0.80/0.30
        & 20.43/0.73/0.35 \\

        AnySplat
        & \xmark & \cmark
        & 23.12/0.80/0.16
        & 25.18/0.83/0.15
        & 741K/3117K
        & 23.18/0.81/0.16
        & 25.42/0.84/0.15 \\

        DepthSplat-B
        & \xmark & \xmark
        & 24.13/0.87/0.19
        & 23.13/0.84/0.22
        & 762K/3175K
        & 23.20/0.88/0.18
        & 24.76/0.90/0.15 \\

        \midrule

        MonoGS
        & \cmark & \cmark
        & 15.15/0.52/0.56
        & 23.90/0.78/0.38
        & 9K/13K
        & 21.17/0.71/0.43
        & 23.23/0.78/0.39 \\

        \midrule

        CUT3R
        & \cmark & \cmark
        & 15.13/0.53/0.57
        & 15.25/0.54/0.57
        & 762K/3175K
        & 15.16/0.51/0.58
        & 14.90/0.50/0.59 \\

        StreamSplat
        & \cmark & \cmark
        & --
        & --
        & 762K/3174K
        & 20.48/0.75/0.33
        & 24.23/0.83/0.26 \\

        Spann3R
        & \cmark & \cmark
        & 18.87/0.60/0.47
        & 19.63/0.62/0.46
        & 762K/3175K
        & 19.07/0.59/0.48
        & 18.50/0.59/0.48 \\

        Point3R
        & \cmark & \cmark
        & 17.36/0.57/0.46
        & 17.01/0.57/0.47
        & 762K/3175K
        & 16.68/0.53/0.48
        & 15.71/0.53/0.51 \\

        StreamVGGT
        & \cmark & \cmark
        & 13.52/0.42/0.65
        & 12.72/0.29/0.72
        & 762K/3175K
        & 11.75/0.32/0.70
        & 11.42/0.23/0.75 \\

        \textbf{S2Tok (Ours)}
        & \cmark & \cmark
        & \textbf{21.23/0.71/0.37}
        & \textbf{20.54/0.71/0.39}
        & \textbf{41K/71K}
        & \textbf{22.50/0.74/0.33}
        & \textbf{22.63/0.76/0.34} \\

        \bottomrule[1.3pt]
    \end{tabular}%
    }
\end{table*}

\input{figures/fig_qual_mipnerf_final.tex}

\input{figures/fig_qual_dl3dv_causal.tex}

\input{figures/fig_qual_scannet_causal.tex}

\input{figures/fig_steps.tex}

\end{document}